\documentclass[11pt]{article}
\usepackage[utf8]{inputenc}

\usepackage[lofdepth,lotdepth,caption=false]{subfig}
\usepackage{fancyhdr}
\usepackage{hyperref}
\usepackage{amsmath, amssymb, amsfonts, graphicx}
\usepackage{xspace}
\usepackage{xfrac}
\usepackage{setspace}
\usepackage[margin=1in]{geometry}

\usepackage{url}
\usepackage{comment}
\usepackage{diagbox}
\usepackage{multirow}
\usepackage{wrapfig}
\usepackage{mathrsfs}
\usepackage{algorithm, algpseudocode}
\usepackage{float}
\usepackage{multicol}
\usepackage{soul} 
\usepackage{xcolor}
\usepackage{makecell}

\newcommand{\be}{\begin{equation}}
\newcommand{\ee}{\end{equation}}

\newcommand{\tq}{\tilde{q}} %
\newcommand{\eps}{{\varepsilon}}
\newcommand{\T}{\mathscr{T}}
\newcommand{\bT}{\mathbb{T}}
\newcommand{\bE}{\mathbb{E}}

\newcommand{\bA}{\mathbf{A}}
\newcommand{\cO}{\mathcal{O}}

\newcommand{\cK}{\mathcal{K}}
\newcommand{\cC}{\mathcal{C}}
\newcommand{\cS}{\mathcal{S}}

\newcommand{\cT}{\mathcal{T}}
\newcommand{\cF}{\mathcal{F}}
\newcommand{\cX}{\mathcal{X}}
\newcommand{\by}{\mathbf{y}}
\newcommand{\hT}{w}

\newtheorem{remark}{Remark}

\title{Continuous-time Optimal Stopping through Deep Reinforcement Learning}
\author{Cosmin Borsa and Mike Ludkovski\thanks{Department of Statistics \& Applied Probability, UC Santa Barbara. Santa Barbara, CA 93106-3110, USA}}
\date{First announced: June 15, 2026}

\begin{document}

\maketitle

\begin{abstract}
Simulation based solvers for optimal stopping problems must discretize the 
stopping decision. Under classical dynamic programming, a coarse exercise grid with only a few stopping opportunities can materially undervalue the optimal expected reward, whereas on a very fine grid, approximation errors accumulate through the backward recursion. To remove this limitation, we develop a new reinforcement-learning inspired algorithm that enables us to learn the exercise rule at arbitrarily fine time resolution. Our CARLOS algorithm utilizes an aggregate deep neural network (ADNN) to learn a joint space-time decision boundary. Starting from a coarse time grid, we progressively increase the frequency of stopping opportunities, while in parallel training the ADNN to refine its timing-value estimates. We moreover design an adaptive sampling strategy that gradually concentrates training effort near the stopping boundary. Benchmarked results show that CARLOS delivers higher prices than  existing Bermudan solvers, approaching the American upper bound, and achieves high computational efficiency relative to to non-RL comparators. 
\end{abstract}

\section{Introduction}

Simulation-based solvers for Optimal Stopping problems (OSP), exemplified by the Longstaff-Schwartz (LSMC) framework \cite{LongstaffSchwartz}, have been a core part of the quantitative finance toolbox over the past 25 years. Indeed, OSPs are ubiquitous, for example arising in the pricing  of all single-name Puts in U.S.~markets. Since these solvers operate with simulated trajectories of the underlying stochastic processes, time discretization is a necessary step in their implementation. Consequently, the convention is to focus on the Bermudan formulation where a pre-specified exercise frequency $\Delta t$ is  given.  

Real-life contracts are American-style and can be exercised at any point. Mathematically, the role of the time-step $\Delta t$ is well understood \cite{DupuisWang} and the LSMC-inspired methods can be implemented at any frequency, so that in theory one can recover the American-style solution at arbitrary precision. In practice, however, LSMC methods have an intrinsic issue of error back-propagation, which tends to be severe. As a result, one is usually limited to quite coarse $\Delta t$'s, creating a material gap to exercise-anytime value. Despite the vast extant literature on machine learning for OSPs \cite{mlOSP}, to our knowledge this gap between feasible Bermudan solvers and the original American specification has never been adequately addressed. 

In this paper we propose a new algorithm that explicitly targets American-style contracts. To this end, we employ Reinforcement Learning (RL) to learn the continuous-exercise optimal stopping strategy.  
We start from a coarse LSMC solver and gradually refine the exercise frequency, employing neural network (NN) surrogates to approximate the underlying timing values. This refinement can be done up to any specified frequency $\Delta t^{ex}$ and because no backward iteration is involved, it essentially eliminates error accumulation. We demonstrate that our method (i) effectively closes the Bermudan-American value gap; (ii) is much more efficient (i.e., faster) than a brute force approach running LSMC at a high exercise frequency. 

The starting idea of our solver is to aggregate the collection of time-discretized stopping rules---indexed by the time step---into a single deep NN surrogate that approximates the entire timing value hypersurface, continuously across $t$ and input state $x$. This NN provides a stopping decision for any $t \in [0,T]$ (rather than on a discrete grid like in traditional solvers), and the main task becomes to train it to learn the continuous-time stopping rule. This shift from the original coarse-grained exercise rule to the desired continuous exercise frequency, raises two novel challenges whose resolution is a key part of our methodological contribution. The first challenge we must contend with is of \emph{concept drift}---the distribution of the training data  changes as training proceeds. This is because the pathwise rewards that underlie the LSMC paradigm are intrinsically tied to the exercise frequency. Hence, even on the same simulated trajectory, the stopping time and the collected reward will shift as exercise frequency is refined. To solve this challenge, in parallel with the RL training, we gradually traverse a collection of exercise grids. In particular, we show that a good rule of thumb is to iteratively \emph{halve} the time steps $\Delta t^{(b)}$ every few RL iterations. 

The second challenge relates to the fact that the stopping region intrinsically shrinks as $\Delta t \to 0$. This implies that some inputs might be in the stopping region for a given $\Delta t$ but end up in the continuation region for a smaller $\Delta t'$. However, the base approach immediately stops any trajectory that is in the stopping region, which effectively prevents re-training the NN to expand the continuation region. To solve this issue, we introduce a novel ``delayed stopping'' technique that adds an exploratory aspect to the training. 

In all, our algorithm, dubbed CARLOS (Continuous-time Adaptive Reinforcement Learning for Optimal Stopping), is initialized via a regular (but coarse) LSMC step and then trains the neural network surrogate via about a dozen RL loops while traversing  3-6 time-discretization levels. The final decision rule treats $t$ as a continuous input and can  be evaluated at arbitrary exercise frequency.

\subsection{Reinforcement Learning for Optimal Stopping}
The dominant approach to pricing of Bermudan options relies on Dynamic Programming (DP) which consists of backward recursion. The LSMC strategy \cite{LongstaffSchwartz} moves backward from option maturity $t_K=T$ to $t_0=0$, using (linear) regression to learn the continuation value which corresponds to a conditional expectation of future expected payoff. Thus, a new regression is done at each time step $t_k$ and is coupled to previous regressions (for $t_\ell>t_k$) that determine those future payoffs, causing error back-propagation. 

The alternative to DP borrows from the Markov Decision Process literature, namely policy- and value-iteration techniques. RL dispenses with the recursive logic and aims to learn the global stopping policy from forward samples. Training data across different time steps is used jointly to improve the approximation in space and time, resembling transfer learning across $t_k$'s.  For optimal stopping, this learning is ``reinforcing'' because the training samples, i.e., the pathwise rewards, are simulated based on the current stopping rule and the stochastic environment. Unlike LSMC, RL therefore seeks a single ``aggregate'' state-action emulator $Q(x,a)$, where state  $s \equiv (t,x)$ now refers both to the stochastic state like underlying asset values, and time $t$. Early versions of such approaches appeared in \cite{tsitsiklis2002optimal,yu2007q,li2009learning} using 
linear approximation (least squares regression against a fixed set of basis functions). Li et al.~\cite{li2009learning} derived bounds on least-squares policy iteration which is the implementation of RL with \emph{linear} representation of action-state map in terms of basis functions. The linear structure allows to express RL policy errors in terms of underlying (finite-sample) projection error. 
 Such Q-learning is the basis for our RL framework.

The special feature of optimal stopping is that the action space is particularly simple, being binary. Denoting by $a=0$ stopping and $a=1$ continuation, the reward from $a=0$ is explicit, so one only needs to model $Q(s,1)$.  
Herrera et al.~\cite{HerreraCalypso} proposed the RRLM variant which uses randomized Q-fitting iterations to learn the state-action emulator.
Another deep-learning inspired implementation of RL for discrete-time optimal stopping is in \cite{LiLee23}.

 One attraction of RL is the ability to handle fully data-driven setups where no such model is available. For American option pricing this corresponds to directly training on past stock trajectories without specifying stochastic dynamics. This ``model free'' idea was explored in \cite{Fathan} and most recently in \cite{damera2023deep,dai2026learning}.  Related control settings where RL is applied include \cite{daluiso2024swing,yang2026synchronizing}.
In our setup, the RL is model-based: the stochastic environment is fully specified and hence arbitrarily many samples can be generated. In particular, we are able to employ adaptive sampling during our training, to preferentially explore regions of interest.

\subsection{Deep Learning for Optimal Stopping}
RL is naturally intertwined with deep learning as policy or value learning go hand in hand with the iterative training of a NN surrogate. Deep learning has been applied extensively to optimal stopping and in this section we summarize the relevant literature.

To our knowledge, the first application of neural networks in LSMC was in Kohler et al.~\cite{Kohler} who employed shallow single-layered NNs to approximate the conditional expectation underlying Snell envelopes. More recently, Lapeyre and Lelong \cite{LapeyreBernard} and Becker et al.~\cite{Becker} considered more advanced deep learning methods, in particular to deal with high-dimensional settings. A key motivation for NNs is to circumvent the well-known challenge of   basis function selection in traditional LSMC. Thus, deep NN emulators are employed as flexible approximators that provably converge \cite{LapeyreBernard,gonon2024deep} (in the regime of increasing the network size) by a suitable variant of the universal approximation theorem. While offering high expressivity, training a NN is a non-convex objective and requires gradient descent iterations. As an intermediate approach between classical least squares regression and NN training, 
Herrera et al.~\cite[Section 2]{HerreraCalypso} proposed the RLSM algorithm where the inner layer weights of the NN are randomly sampled and only the last layer is optimized. This randomized approach allows to retain a convex objective, solved via classical linear regression equations, and can be understood as picking expressive random bases. 

All the above works maintain the DP logic of backward recursion, constructing a separate NN emulator at each $t_k$. In practice, these emulators are very similar, since  the stopping policies at two adjacent time steps are so. This observation is not new; for example both \cite{LapeyreBernard} and \cite{Becker} exploit it during backward recursion, by re-using the same NN object and gradually updating it. Such warm starts leverage the gradient descent paradigm of deep learning and substantially speed up training time. Taking this logic a step further,
Guo et al.~\cite{Langrene} proposed a single NN for $Q(s,1)$ that takes time  $t$ and location $x$ as inputs to approximate continuation values across space-time. This variation increases the prediction accuracy  while decreasing the computational time.  To train their NN, \cite{Langrene} initially set the stopping time to maturity and perform training-updating loops in a basic reinforcement learning setting until a predefined criterion is met. 

Additional NN-based approaches were proposed by \cite{Cheridito} who targeted learning the 0/1 stopping decision rule and by  \cite{ValentinTissot} who approximated the epigraph of the stopping set.  Finally, we mention the body of work that utilizes NNs to solve a free-boundary partial differential equation for the option price derived via Feynman Kac formulas, see e.g.~deep Galerkin methods \cite{sirignano2018dgm} and backward stochastic differential equations \cite{chen2021deep,gao2023convergence,yang2024deep} both of which specifically tackle American option pricing. 

From the implementation side, the precise architecture of the NN can make a significant difference. Dense feed-forward networks have been used in \cite{Kohler} (a shallow 1-layer version), as well as \cite{HerreraCalypso,LapeyreBernard,Becker}. \cite{Felizardo} proposed convolutional NNs, while \cite{damera2023deep} proposed recurrent NNs. For RL-type methods, \cite{Fathan} applied customized Double Deep Q-Network (DDQN), Categorical Distributional RL, and Implicit Quantile Networks, deploying
LSTM architecture with a dynamic layer and a dropout wrapper to capture long-term dependencies in sequential data. 
\cite{Ery} employ a fully connected feed-forward NN for a Q-learning approach to recover the optimal stopping times. 

One of the motivations for our study is to control error back-propagation in LSMC. In that vein, we also mention various modified LSMC methods \cite{Zhang,Shevchenko} that correct the errors accumulated during the backward iteration steps.

The rest of the paper is organized as follows. Section \ref{sec:OS-via-DP} sets up the simulation-based framework for solving optimal stopping problems. Section \ref{sec:New-Algo} presents our new CARLOS algorithm. After a couple of illustrations with a classical 1-dim Put and 2-dim Max Call options, in Section \ref{sec:Results} we present the benchmarked results over a collection of American options that have been considered in the literature.  Section \ref{sec:RL-details} provides the full methodology of our method, including input selection, output generation, and neural network construction details. Along the way, we discuss the key tuning parameters of CARLOS through several comparative statics experiments and gives a guidance for users. Section \ref{sec:Conclude} concludes.

\section{Optimal Stopping via Dynamic Programming}\label{sec:OS-via-DP}

We adopt a state-space framework: let $(X_t)$ be  
the $d$-dimensional Markov stochastic state process on a probability space $(\Omega, \mathbb{P}, \cF)$, taking values in $\mathcal{X} \subset \mathbb{R}^d$ and adapted to a filtration $\mathbb{F} :=(\mathcal{F}_{t})_{t \in [0, T]}$. The reward at time $t$ is given by $h(X_t)$, where $h:  \mathcal{X} \mapsto \mathbb{R}$ satisfies 
\begin{equation}
    \label{eq:OS1} 
	\mathbb{E}\big[\sup_{0 \le t \le T}|h( X_{t})|\big] < \infty.
\end{equation}
We interpret $h( x)$ as the (time-stationary) payoff obtained if the state is $x \in \cX$, and $\mathbb{P}$ as the pricing (\emph{risk-neutral}) measure. For instance, for a Put option $h(x) =  (K-x)_+$.
Let $r$ represent the discount rate, assumed to be constant; stochastic discounting or time-dependent payoffs can be embedded as a coordinate of $X$ and then subsumed in $h(\cdot)$. 

Consider a contract maturity $T$.
We use $\bT$ to denote a \emph{time grid}; in what follows for simplicity we restrict attention to uniform grids with a constant step. Indexing grids by their size $N$, we write $\bT^{(N)} = \{t_0=0, t_1, \ldots, t_N=T\}$ with 
\begin{equation}
    \label{eq:OS1}
    \Delta t^{(N)} = t_{n+1}-t_n = \frac{T}{N}, \qquad n=0,\ldots,N-1.
\end{equation}
Let $\cT_t$ denote the set of all $\mathbb{F}$-stopping times taking values in $[t,T]$, and similarly $\cT^{(N)}_n$ for the $\bT^{(N)}$-valued stopping times that take values in $\tau \in \{t_n, t_{n+1}. \ldots, t_N\}$. When $N$ is clear we drop the superscript and just write $ \cT_n$. 

The Bermudan formulation of the OSP for a given frequency $\Delta t^{(N)} = T/N$ is to  compute the value function $V^{(N)}: \bT^{(N)} \times \mathcal{X} \mapsto \mathbb{R}$, 
\begin{equation}
    \label{eq:OS2} 
	V^{(N)}(t_n, x) := \sup_{\tau \in \mathcal{T}^{(N)}_n} \mathbb{E} \left[e^{-r (\tau - t_n)} h( X_{\tau})|\ X_{t_n} = x \right], \qquad x \in \cX, \ n \in \{0, \ldots, N-1\},
\end{equation}
by finding an optimal stopping time $\tau_n^{\star} \in \mathcal{T}^{(N)}_n$ at which the supremum of the discounted reward process $\{ e^{-rt_k}h( X_{t_k})\}_{k=n}^N$ is attained \cite[p.~12]{PeskirBook}.
The American formulation is to solve for
\begin{align}
V(t, x) := \sup_{\tau \in \mathcal{T}_t} \mathbb{E} \left[e^{-r (\tau - t)} h( X_{\tau}) |\ X_{t} = x \right].
\end{align}

Classical results \cite{DupuisWang} imply that $V^{(N)}(t,x) \uparrow V(t,x)$ as $N \to \infty$, with a difference on the order of $\cO(N^{-1})$. Hence, a sufficiently fine-grained Bermudan option can be used to approximate arbitrarily well the value of the American contract.

\subsection{Backward Learning}\label{sec:LSMC}
In this section we fix the grid $\bT^{(N)}$ and review the sequential LSMC approach for approximating $V^{(N)}(0,x_0)$. This is both to contrast with RL and because LSMC is a way to initialize our overall algorithm.
The idea of LSMC is to approximate the optimal stopping strategy on $\bT^{(N)}$  by pathwise \textbf{dynamic programming}: deriving a sequence of stopping times $\tau_n^\star$, indexed by the time step $t_n$, by backward induction from  $t_N=T$. 
We focus on the  \textit{timing values} ${\T}(t_n,x)$ defined as the difference between the continuation value $\tilde{q}(t,x)$  and the immediate reward $h(x)$:
\begin{align}
    \label{eq:LSMC-TV} 
    {\T}(t_n, x) & := \tq(t_n, x) - h( x) \qquad \text{where} \\
     \label{eq:LSMC1} 
    \tq(t_n, X_{t_n}) & := \bE \left[ e^{-r(\tau_{n}^{\star} -t_n)} h( X_{\tau_{n}^{\star}}) \mid  \cF_{t_n} \right],
\end{align}
where $\tau_{n}^{\star}$ is an optimal stopping time in $\mathcal{T}_{n}$. The theory of Snell envelopes identifies $\tau_n^\star$ with the first instance when the timing value becomes non-positive, 
\begin{equation}
    \label{eq:LSMC2} 
    \tau_{n}^{\star} = \min\left\{t_k \in \cT_{n} : \tq(t_k, X_{t_k}) \le h( X_{t_k})\right\} = \min\left\{t_k \in \cT_{n} : {\T}(t_k, X_{t_k}) \le 0\right\}.
\end{equation}

Using \begin{align}\label{eq:LSMC3}
    \tq(t_n, X_{t_n})  
    &= e^{-r\Delta t}\mathbb{E} \left[\tq(t_{n+1}, X_{t_{n+1}})\mid \cF_{t_n} \right],
\end{align}
one-step-ahead pathwise samples of $\T(t_n,\cdot)$ are obtained via 
\begin{equation}
    \label{eq:LSMC-TV-path} 
    \hT(t_n, x_{n}^k) := e^{-r\Delta t}\tq(t_{n+1}, x_{{n+1}}^k) - h( x_{n}^k),
\end{equation}
where $x_{{n+1}}^k$ is a sample of $X_{t_{n+1}}$ conditional on $X_{t_n} = x^k_{n}$, for a collection $k \in \{1, \ldots, K\}$.  The typical construction is to create $K$ trajectories of the process $X$ labeled cross-sectionally as $\mathbf{x}_{n}= \{x^k_{n}, k=1,\ldots,K\}$, for $n=0,\ldots, N$.

One may lift from these samples to the overall $L^2$-prediction by
regressing the $K$ pathwise timing values $\hT(t_{n}, \mathbf{x}_{{n}})$ against $\mathbf{x}_{n}$. The least-squares regression yields a surrogate $R_n$ whose predicted timing values $\hat{\T}(t_n, \mathbf{x}_{n})$ define the pathwise stopping times $\{\tau_{n}^k\}$ and pathwise continuation values $q^k_n \equiv \tq(t_n, x_{n}^k)$ as follows
\begin{equation}
    \label{eq:LSM4} 
    (\tau_n^k, q^k_n) = \begin{cases}
    (t_n, h( x_{n}^k)) &\text{if } \hat{\T}(t_n, x_{n}^k) \le 0 \text{ and } h(x_{n}^k) > 0;\\
    (\tau_{n+1}^k, e^{-r\Delta t}q^k_{n+1}) &\text{otherwise}.
    \end{cases}
\end{equation}
The recursion for $q_n^k$ is based upon $\tq(t_n, x_{t_n}^k) \simeq e^{-r(\tau_n^k - t_n)} h( x_{\tau_n^k}^k)$, where the latter expression is an unbiased estimator of \eqref{eq:LSMC1}. Another familiar functional representation is $$\tq(t_n, x) = \max\left( \mathbb{E}[e^{-r \Delta t} \tq(t_{n+1},X_{t_{n+1}})|X_{t_n}=x], h(x) \right).$$

For a given trajectory $\{x_{{n}}^k\}_{n=0}^N$, if the immediate payoff at a time step $t_n$ is positive and the estimated timing value is non-positive, then we should \textit{stop} $\tau_n^k = t_n$. Otherwise, we should \textit{continue}, and the pathwise stopping time is the one computed in the time step $\tau_{n+1}^k$. For a fixed step $n$, the set $(\tau_n^k)_{k = 1}^K$ can be interpreted as a sample of $\tau_n^{\star}$. 
Finally, the backward recursion is run from $n=N$ down to $n=0$ to yield a sequence of regression emulators $\{R_n\}_{n=0}^{N-1}$. 

\begin{remark}
We use neural networks as the regression emulators $\{R_n\}_{n=1}^{N-1}$, training them via stochastic gradient descent. 
 The ``first'' network $R_{N-1}$ is initialized with random parameters $\tilde{\theta}_{N-1}$, drawn from Gaussian or continuous uniform distributions \cite{Kohler}.
 For subsequent networks, to improve performance their weights and biases are initialized by an average of  the previous (in $t$) weights and biases, cf.~\cite{Langrene} since the current parameters $\tilde{\theta}_n$ is similar to the previous $\tilde{\theta}_{n+1}$. This warm starting reduces training time.
\end{remark}

The timing values are converted into a  stopping rule $\phi$ according to
\begin{equation}
    \label{eq:Stop-Decision} 
    \phi(t, x) = 
    \begin{cases}
        1 & \text{if } \hat{\T}(t, x) \le 0 \text{ and } h( x) \ge 0 \text{, or if } t = T, \\
        0 & \text{otherwise};
    \end{cases}
\end{equation}
where $\phi(t,x)=1$ indicates stopping and $\phi(t,x)=0$ continuation. Globally, the stopping strategy is summarized by the stopping region $\cS := \{ (t,x): \phi(t,x) = 0\}$. To obtain the ultimate option price, a separate forward evaluation Monte Carlo is needed. For a path $\{x_t\}$ starting from arbitrary $(t_{init}, x_{init})$ and progressing along the exercise grid $\bT$, the payoff is $h( x_{\tau})$, where $\tau$ is the first time it leaves the continuation region $\cC := \{ (t,x) : \phi(t,x) = 1 \}$,
\begin{equation}
    \label{eq:First-Time} 
    \tau := \min\{t \in \{ t_{\text{init}}\} \cup \bT : \phi(t, x_t) = 0\} \wedge T.
\end{equation}
Expected payoff, aka option price with initial state $x_0$, is obtained as an empirical average over $M$ paths originating at $(0, x_0)$.

\subsection{Time Discretization Effects}\label{subsec:LSMC-Errors}
Before explaining CARLOS, we discuss the dual role of time stepping, namely for decision optimization and for forward evaluation. 

The LSMC paradigm generates two fundamental sources of error. First, it works with stopping rules taking values in $\bT^{(N)}$ rather than in continuous time. Second, the training of the surrogates $R_n$ is iterative in $n=N,\ldots, 0$ and errors back-propagate. Further numerical errors, such as function approximation error (how well can the surrogate approximate the true $\T$) and Monte Carlo noise (how much do the particular samples affect the fitted network) can be in principle removed as sufficiently wide/deep NNs  are arbitrarily expressive, and sufficiently many stochastic gradient descent steps make the Monte Carlo noise negligible.

In contrast, the discretization error directly conflicts with the back-propagation. If we make $N$ larger ($\Delta t$ smaller), then there is more error accumulation. This issue is well known in the folklore, and nearly all benchmarks include just a few time periods. Indeed, the typical recommendation is to keep the number of steps $N$ well below 100, which may be reasonable for $T<1/2$, but otherwise becomes quite restrictive.  For example, the oft-cited benchmarks in \cite{AndersenBroadie} use $T=3, N=9$, meaning there are just 9 exercise opportunities, with stopping allowed once every 4 months ($\Delta t^{(N)} = 1/3$), an extreme example of Bermudization.

There are in fact two different discretization parameters when using a discrete-time solver for OSP. The exercise frequency $\Delta t^{ex}$ controls when the reward may be collected during forward evaluation. Large $\Delta t^{ex}$ values correspond to fewer stopping opportunities. The solver or ``training'' frequency $\Delta t^{tr}$ determines how often the maximum operator is applied when computing the Snell envelope.
This parameter controls the resulting stopping policy, i.e.,~the exercise rule. While classically $\Delta t^{ex}=\Delta t^{tr}$, they can be different: for instance, one may allow stopping very frequently even if the stopping policy is obtained from a coarse time grid; or conversely, one might have a fine-grained policy but exercise only rarely. 

Literally taking $\Delta t^{ex} \to 0$ is unnecessary. First, the loss when using $\Delta t^{ex} = T/N$ compared to continuous stopping when computing timing values is theoretically $\cO(N^{-1})$ \cite{DupuisWang}, so there is limited upside to refining ad infinitum. Second, to evaluate a rule with exercise frequency $\Delta t^{ex}$ takes effectively $\cO( (\Delta t^{ex})^{-1}) = \cO(N)$ effort since we must evaluate the machine learning surrogates $R_n(\cdot)$ for every $n=1,\ldots,N$ (doubling the number of steps will require double the evaluations of $R_n$'s). Third, the learning becomes more challenging as $\Delta t^{ex} \to 0$ due to the smooth fit principle: in discrete time, there is (theoretically) a crisp classification between a timing value being positive (and hence not stopping) or negative. However, in continuous time, we have $\partial_x \tilde{\T}(t, x) = 0$ and $\min_x \tilde{\T}(t, x) = 0$ so that the value function smooth pastes to the payoff function and the timing value is at least zero, never negative. This makes it harder to numerically determine whether to stop or continue at an input~$(t,x)$ close to the boundary.

\begin{figure}[!ht]
    \centering
    \includegraphics[width=0.485\textwidth,trim=0.2in 0.2in 0.2in 0.2in]{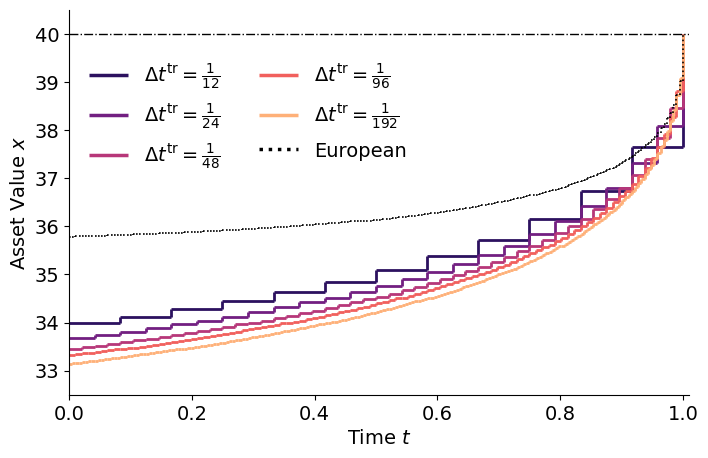}
    \caption{Optimal stopping boundaries of the 1-dimensional Bermudan Put option \texttt{B1} from Table~\ref{tab:contracts-tab} computed via the CN method at various solver frequencies~$\Delta t^{tr}$. These boundaries were obtained by converting $\phi(t,x)$ into contour lines. The CN method used $\Delta t^{\text{PDE}} = 1/192$ and $\Delta x = 0.02$. The dashed contour shows the thresholds $B^{[0]}$ where the European Put becomes less valuable than its intrinsic value, cf.~\eqref{eq:EU-BD}.}    
    \label{fig:BP1-Boundaries} 
\end{figure}

To illustrate, we price a 1-dimensional Bermudan Put option (\texttt{B1} from Table~\ref{tab:contracts-tab} below) using a variety of exercise frequencies $\Delta t^{ex}$ and solver frequencies $\Delta t^{tr}$.  For the latter, we use the Crank-Nicolson partial differential equation solver (CN, which is second-order accurate in both time and space and is unconditionally stable \cite[p.~156]{Wilmott})\footnote{Both the CN and explicit finite-difference schemes use a time step $\Delta t^{PDE} \le \Delta t^{tr}$ to approximate time derivatives.  The time step $\Delta t^{PDE}$ is inherent to the finite-difference method itself and chosen independently.}. Figure \ref{fig:BP1-Boundaries} displays the optimal exercise boundaries  under various $\Delta t^{tr}$'s. Each boundary $B^{[I]}$ is a right-continuous  function with steps at each $\{i\Delta t^{tr}: i = 0, 1, \ldots, I\}$, where $I = T/\Delta t^{tr}$.   A larger solver frequency $\Delta t^{tr}$ lowers potential future gains (as there are fewer opportunities to stop) and therefore makes the controller more impatient about collecting the reward early. Accordingly, in Figure \ref{fig:BP1-Boundaries} as $\Delta t^{tr}$ increases, the boundaries shift upwards.  We note that as $\Delta t^{tr}$ decreases, the boundaries $B^{[I]}$ rapidly converge (to the boundary $B^{[\infty]}$ of the American formulation). At the other extreme, we plot the \textit{European boundary} $B^{[0]}$ that delineates the region 
\begin{equation}
    \label{eq:EU-BD} 
	\big\{(t, x) \in [0, T] \times \mathcal{X}: \mathbb{E}[e^{-r(T - t)} h( X_T) \big|\, X_t =x] = h( x)\big\}.
\end{equation}
where the immediate payoff $h( x)$ exceeds the price of the European-style contract, effectively corresponding to the Bermudan formulation with $\Delta t^{tr} = T$. The European boundary delineates the region where stopping decision must be evaluated---above it continuation is a priori optimal.

Table~\ref{tab:PDE-tab} shows how the interplay between the exercise frequency $\Delta t^{ex}$ and the solver frequency $\Delta t^{tr}$ influences the expected reward. 
The classical case of $\Delta t^{ex}=\Delta t^{tr}$ is on the diagonal and shows that as expected more frequent stopping opportunities capture greater payoffs. The more interesting behavior is off-diagonal.  When $\Delta t^{ex} > \Delta t^{tr}$, infrequent stopping ``leaves money on the table'' by skipping exercise opportunities. This results in lower rewards compared to  $\Delta t^{tr} = \Delta t^{ex}$. 
Conversely, when $\Delta t^{ex} < \Delta t^{tr}$, the stopping rule is too cautious: e.g., looking at the blue boundary in Figure \ref{fig:BP1-Boundaries} when the red one is optimal, yet the ability to stop more frequently still contributes to higher rewards. Making  $\Delta t^{ex}$ extremely small  may lead to lower rewards. Testing the statistical significance of price differences between adjacent $\Delta t^{ex}$ values in Table \ref{tab:PDE-tab} suggests that for a given $\Delta t^{tr}$ highest rewards are obtained when $\Delta t^{ex} \simeq 1/4 \times \Delta t^{tr}$\footnote{These findings are consistent across contracts, see Table \ref{tab:PDE-tab-app} in the Appendix.}. Finally we note that the MC-based payoffs are slightly different from those reported by the PDE solver due to the finite-difference and finite-domain boundary errors. 

\begin{table}[ht]
\centering
\begin{tabular}{c|rrrrr|r}
    \hline
     {\diagbox{$\Delta t^{tr}$}{$\Delta t^{ex}$}} &    
         $\frac{1}{12}$ & $\frac{1}{24}$ & $\frac{1}{48}$ & $\frac{1}{96}$ & $\frac{1}{192}$ &{PDE} \\
    \hline\hline
    $\frac{1}{12}$ & 4.571 & 4.584 & \textbf{4.587} & 4.585 & 4.582 & 4.567 \\
      $\frac{1}{24}$ & 4.568 & 4.586 & 4.593 & {$\mathbf{4.594}^{**}$} & $4.593^{**}$ & 4.582 \\
      $\frac{1}{48}$ & 4.564 & 4.585 & 4.594 & \textbf{4.597} & $4.597^{**}$ & 4.589 \\
      $\frac{1}{96}$ & 4.559 & 4.583 & 4.594 & 4.598 & \textbf{4.600} & 4.593 \\
      $\frac{1}{192}$ & 4.553 & 4.579 & 4.592 & 4.599 & \textbf{4.601} & 4.595 \\
    \hline
\end{tabular}
\caption{Average payoffs for the \texttt{B1} contract from Table~\ref{tab:contracts-tab}, estimated using $1.6 \times 10^6$ Monte Carlo simulated paths and PDE-based stopping rules. Highest value in each row is bolded. Row-wise price differences between adjacent $\Delta t^{ex}$ values are evaluated for statistical significance; standard deviations are approx.~0.0025 throughout. If the difference between a price and its right neighbor is not statistically significant, the right-side price is marked with $\,^{*}$ at the 0.05 level and $\,^{**}$ at the 0.01 level.  For reference, the PDE-based prices for each $\Delta t^{tr}$ are also provided in the last column; the respective CN parameters are $\Delta t^{\text{PDE}} = 1/192$, $\Delta x = 0.02$.
\label{tab:PDE-tab}} 
\end{table}

Table \ref{tab:PDE-tab} is one of the motivations for our RL approach. Indeed, we wish to ``traverse'' the table from the top-left (coarse Bermudan-style optionality) to the bottom-right (fine-grained American-style), by gradually refining $\Delta t^{ex}$ and $\Delta t^{tr}$. The Table shows that we need not to proceed strictly along the diagonal and crystallizes the distinction between training and forward-evaluation discretizations. Part of our objective is then to analyze how much $\Delta t^{tr}$ must be reduced to obtain a sufficiently accurate American option price estimate and how to progressively maximize the extracted payoff as exercise grids are refined.

\section{New Algorithm}\label{sec:New-Algo} 

\subsection{Reinforcement Learning}\label{subsec:RL-Basic}

To overcome the limitations of the classical LSMC method we develop a new RL-driven two-step approach: 
\begin{enumerate} 
\item[Stage 1a]: Run the classical LSMC algorithm to construct a sequence of regression emulators. 
\item[Stage 1b]: Train an aggregate deep neural network using the data from Stage~1a;
\item[Stage 2]: Run an RL algorithm that gradually refines the timing values by retraining the ADNN.
\end{enumerate}

In Stage~1, we use a coarse time grid and aim for a  moderate accuracy level.
The goal is to have a reasonable starting point for the RL:  the primary effort and computational resources are reserved for Stage~2. 
To initialize the RL, we stack the training data $(\mathbf{y}, (\mathbf{t}, \mathbf{x}))$ generated by the classical LSMC  in Stage~1a across all the time steps and use it to train a global regressor $R^{\Theta}$, henceforth referred to as the \textbf{aggregate deep neural network} (ADNN), where $\Theta$ denotes its trainable parameters (distinguished from the NN parameters $\theta_n, n=0,\ldots, N$ in Stage~1a). The ADNN incorporates both the time and spatial dimensions as inputs, setting the stage to learning out-of-sample timing values corresponding to smaller solver frequencies.

The core RL iterations in Stage~2, 
indexed by the loop counter  $\ell$, serve two concurrent objectives. The first role of the loop is to refine the stopping frequency $\Delta t^{tr}$. The second role is to converge to the optimal ADNN weights through training on additional inputs, resolving inaccuracies inherited from Stage~1 while mitigating error accumulation. 
The two objectives must be achieved gradually and in parallel due to the underlying challenge of \textit{concept drift}: changing the distribution of training inputs and outputs. 

To formalize our grid refinement, we define a sequence of nested time grids $\bT^{(b)} \supseteq \bT^{(b -1)}$, $b = 0, \ldots, B$ and label the exercise grid as $\bT^{(ex, b)}$ and the solver grid as $\bT^{(tr, b)}$. 
We use the coarsest $\bT^{(tr, 0)}$ for Stage~1 which also serves to initialize the RL in Stage~2.  

To update the ADNN $R^{[\ell]}$,  in each loop $\ell$ we generate a set of $M$ input-output pairs  $(\mathbf{y}, (\mathbf{t}, \mathbf{x}))^{[\ell]}$, with boldface denoting vectors $\by^{[\ell]} =y^{1:M}$, etc.  A detailed explanation of how these input-output collections are constructed is provided in Sections~\ref{subsec:RL-Inputs} and~\ref{subsec:RL-Outputs}. 
After each training update, algorithm logic is used to decide what should be the next exercise frequency: either keeping the same grid $\bT^{(tr,b)}$ or shifting to the next finer grid $\bT^{(tr,b+1)}$, see Section \ref{sec:refine-grid}.
The learning continues until the ADNN has been trained on the finest solver grid $\bT^{(tr, B)}$, see the pseudocode of CARLOS in Algorithm [\ref{alg:RL-Basic}].

\begin{algorithm}[!ht]
\caption{CARLOS: Continuous-time Adaptive Reinforcement Learning for Optimal Stopping}\label{alg:RL-Basic} 
\begin{algorithmic}[1]
\Require Initial ADNN $R^{[0]}$; \textit{RL parameters} e.g. Solver grids $\bT^{(tr, B)} \supseteq \ldots \supseteq \bT^{(tr, 0)}$, Learning rate $\eta^{[0]}$, Training set size $M$; \textit{Path and contract parameters}.
\State Initialize loop count $\ell \gets 0$ and grid index $b \gets 0$
\While{$b \leq B$}
    \State Generate $M$ inputs $(\mathbf{t}, \mathbf{x})^{[\ell]}$ 
    \State Generate $M$ respective outputs $\by^{[\ell]}$ based on rewards on paths with step-size $\Delta t^{(ex),b}$
    \State Update to ADNN $R^{[\ell+1]}$ by training $R^{[\ell]}$ on $(\by, (\mathbf{t}, \mathbf{x}))^{[\ell]}$ with learning rate $\eta^{[\ell]}$
    \State Set $\ell \gets \ell+1$
    \State Update grid level $b$ and learning rate $\eta^{[\ell+1]}$
\EndWhile
\State\Return{Final refined ADNN $R^{[\ell]}$}
\end{algorithmic}
\end{algorithm}

For visualization purposes, we primarily track the stopping boundaries associated with each $R^{[\ell]}$. Given the ADNN $R^{[\ell]}$, the corresponding stopping boundary is defined as
\begin{equation}
    \label{eq:ADNN-BD} 
	B^{[\ell]} =\big\{(t, x) \in [0, T] \times \mathcal{X}: \hat{\T}^{[\ell]}( x) = 0\big\}
\end{equation}
where $\hat{\T}^{[\ell]}$ denotes the timing value estimated by $R^{[\ell]}$. The boundary separates the stopping region $\cS^{[\ell]}$ and the continuation region $\cC^{[\ell]}$.

\subsection{Time Grid Refinement}\label{subsec:RL-Time}

The ADNN $R^{[\ell]}$ is trained on input-output samples which directly depend on the underlying solver frequency $\Delta t^{tr, b}$. Consequently, if one were to initialize the ADNN with a coarse solver grid $\bT^{(tr, 0)}$ and then immediately switch to a fine time grid $\bT^{(tr, B)}$, the underlying training samples would shift significantly, strongly affecting the next learning steps. The mismatch between what the network has learned so far and what it is shown next creates an extrapolation error that hurts learning. To give the ADNN time to adapt, we therefore gradually refine the time grids so that we simultaneously learn to stop on a finer grid and discover the best ADNN parameters $\Theta$ governing that strategy. 

At each learning loop $\ell$, CARLOS adaptively decides whether to transition to a finer solver grid or keep it the same. To this end, we assess incremental gains via a set of \emph{validation paths}. Namely we fix a database of $V$ paths $\{\mathbf{x}\}^{1:V}$ and compute the reward vector $\Upsilon^{[\ell]}$ over these $V$ paths $\{\mathbf{x}\}^{1:V}$ using the grid $\bT^{(ex, b)}$ and the ADNN $R^{[\ell]}$. The pathwise reward differences  $\mathbf{D}^{[\ell]} := \Upsilon^{[\ell]}-\Upsilon^{[\ell-1]}$ across successive loops are used to ascertain whether the learning at level $b$ has saturated. 
Once average $\mathbf{D}^{[\ell]}$ is not statistically different from zero, 
the RL algorithm advances to the next grid~$\bT^{(tr, b+1)}$ in the schedule, see Section \ref{sec:refine-grid}. 

\begin{figure}
    \centering 
    \includegraphics[width=0.485\textwidth,trim=0.1in 0.4in 0.1in 0in]{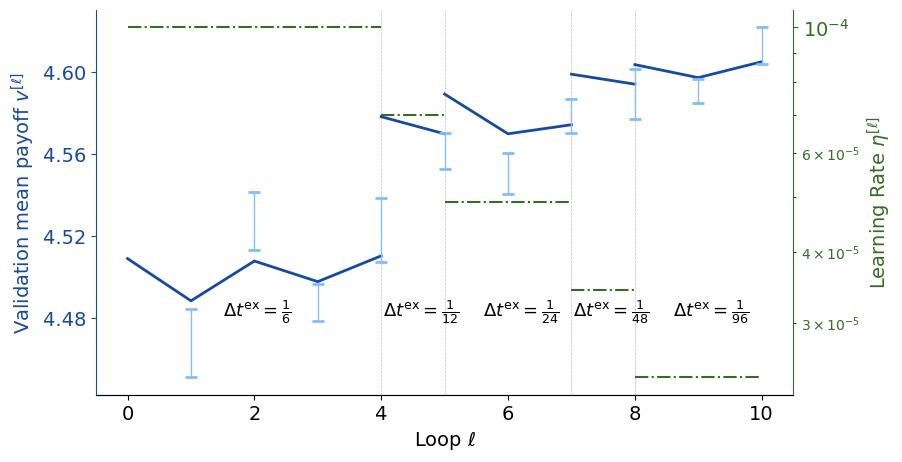}
     \includegraphics[width=0.485\textwidth,trim=0.1in 0.25in 0.4in 0.25in]{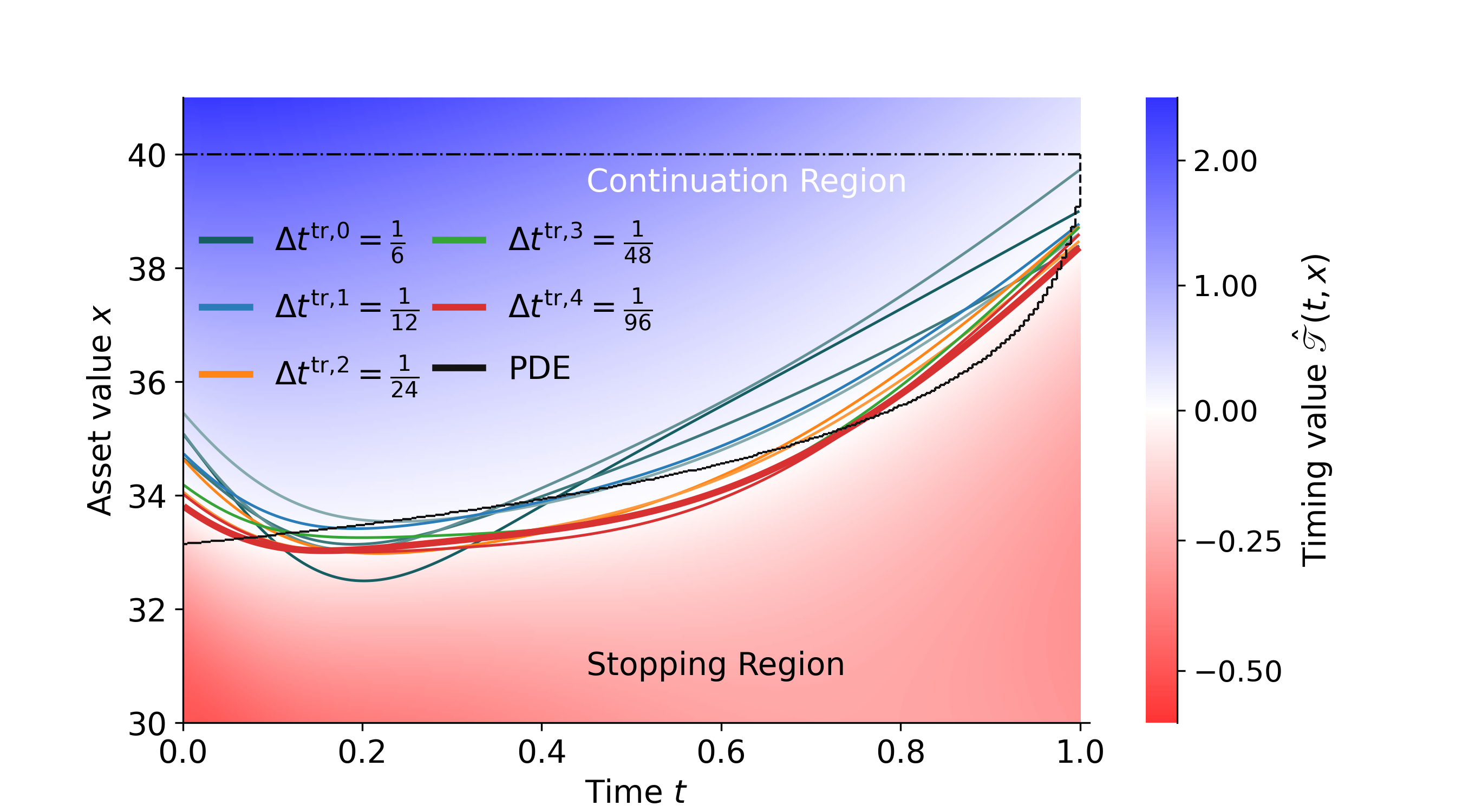}
    \caption{\emph{Left Panel:} Average rewards $\upsilon^{[\ell]} = Ave(\Upsilon^{[\ell]})$ (left y-axis) and learning rates $\eta^{[\ell]}$ (right y-axis, log scale) across RL loops $\ell$ and exercise grids $\bT^{(ex, b)}$ for the \texttt{B1} option from Table~\ref{tab:contracts-tab}. Stopping intervals associated with reward differences $\mathbf{D}^{[\ell]}$ are vertically shifted by $\upsilon^{[\ell]}$ and displayed as blue bands. 
    \emph{Right Panel:} Stopping boundaries (obtained using a marching-squares algorithm) $B^{[\ell]}$, $\ell=0,\ldots, L$, after each RL iteration as the step size is refined from $\Delta t^{tr,0} = \frac{1}{6}$ to $\Delta t^{tr,4} = \frac{1}{96}$. Distinct colors indicate different grids $b$, with progressively lighter shades marking repeated loops on the same time grid $\bT^{(tr,b)}$. The dashed line demarcates the out-of-the-money region. The heatmap displays the final  timing value $\hat{\mathscr{T}}^{[L]}(\cdot, \cdot)$ at loop $L = 10$, with its zero-contour corresponding to the final stopping boundary drawn with a thicker line. Parameter configuration is reported in Table~\ref{tab:Parameter-Setting}.
    }
    \label{fig:B1}
\end{figure}

When transitioning to a new solver grid, we decrease the learning rate according to $\eta^{[b+1]} = \alpha_{\text{dec}} \cdot \eta^{[b]}$. This stabilizes training and preserves recent performance gains by limiting further changes to the ADNN’s learned parameters. We take $\eta^{[0]} = 10^{-4}$ and $\alpha_{\text{dec}} = 0.7$ as the default initial learning rates and decay factors. 
The RL ends when the ADNN has been fully trained on the finest grid $\bT^{(tr, B)}$. 

\begin{remark}
The RL algorithm may also be terminated when the reward difference between consecutive solver grids, $\bT^{(tr, b)}$ and $\bT^{(tr, b-1)}$, is statistically insignificant. Specifically, one can test the reward difference $\Upsilon^{[\ell]} - \Upsilon^{[\ell-1]}$ using the ideas of Section \ref{sec:refine-grid}
\end{remark}

The left panel of Figure~\ref{fig:B1} illustrates the progression of the average rewards $\upsilon^{[\ell]}:=Ave(\Upsilon^{[\ell]})$ as a function of $\ell$ for the \texttt{B1} contract in Table~\ref{tab:contracts-tab}. While $\upsilon^{[\ell]}$ tends to increase, the pattern is not monotone. Moreover, jumps occur when a finer exercise grid is selected. The right y-axis of the figure displays the corresponding learning rates $\eta^{[\ell]}$ which decrease as $\Delta t^{tr}$ decreases. 

The right panel of Figure~\ref{fig:B1} shows the evolution of the stopping boundaries $B^{[\ell]}$'s of this Bermudan Put \texttt{B1} across the successive $\ell=1,2,\ldots, L = 10$ RL loops. As expected, the continuation region expands as the time grid is refined, so that the stopping boundaries shift ``downward'' and also get steeper near maturity $T$. However, this expansion is not monotone: in loop 5, the RL overshoots because the ADNN mis-estimates timing values near maturity $T$ on a solver grid with step size $\Delta t^{tr} = \tfrac{1}{48}$; once the grid is further refined in loop 6 to $\Delta t^{tr} = \tfrac{1}{96}$, the boundary shifts back upward. This demonstrates that the CARLOS algorithm can correct earlier errors and recover boundaries that yield higher rewards, as evidenced in the left panel of Figure~\ref{fig:B1}.  

\begin{figure}[ht]
    \centering
        \includegraphics[width=0.475\textwidth,trim=0.25in 0.2in 0.15in 0.15in]{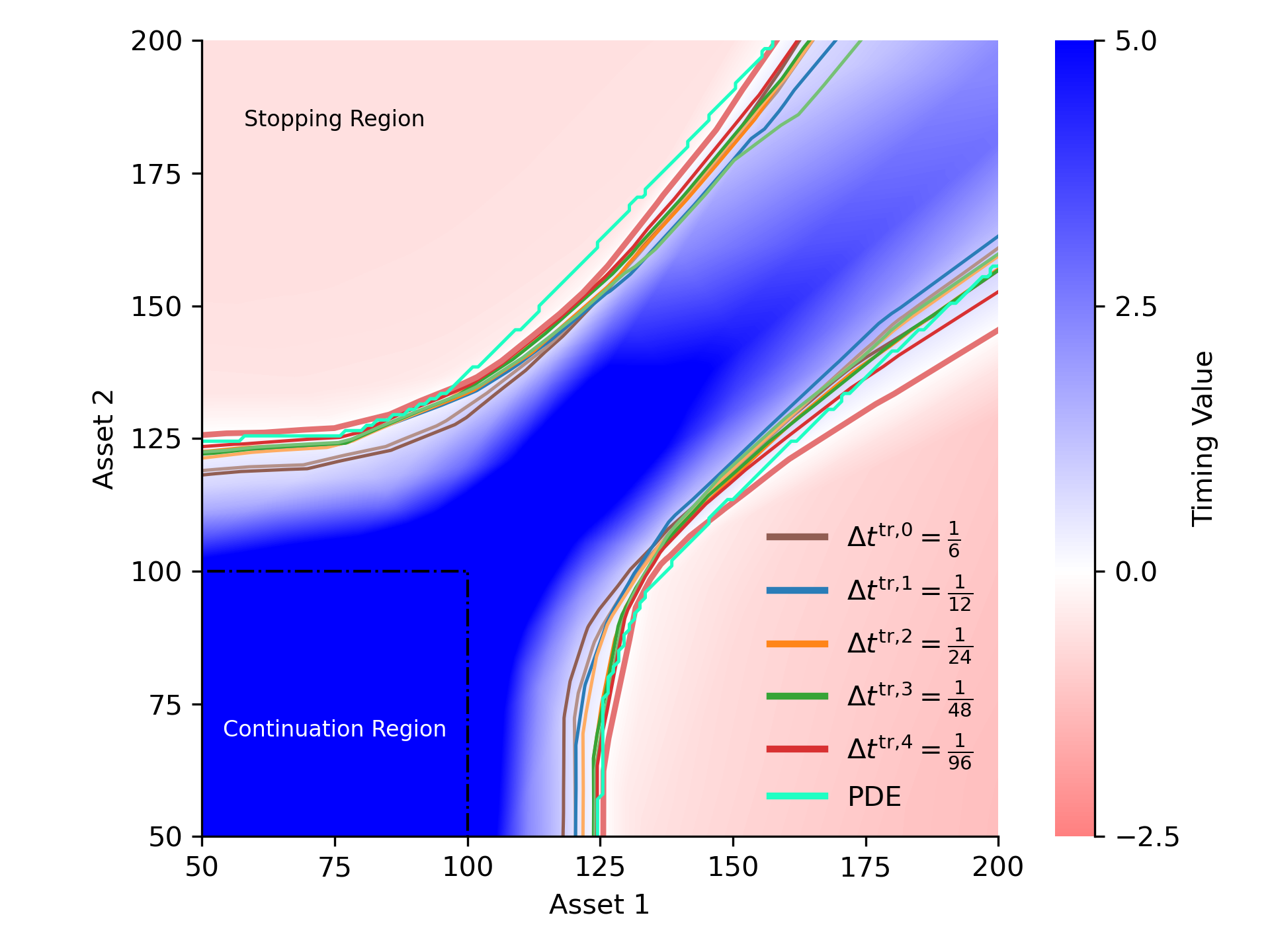}
        \includegraphics[width=0.475\textwidth,trim=0.15in 0.2in 0.2in 0.15in]{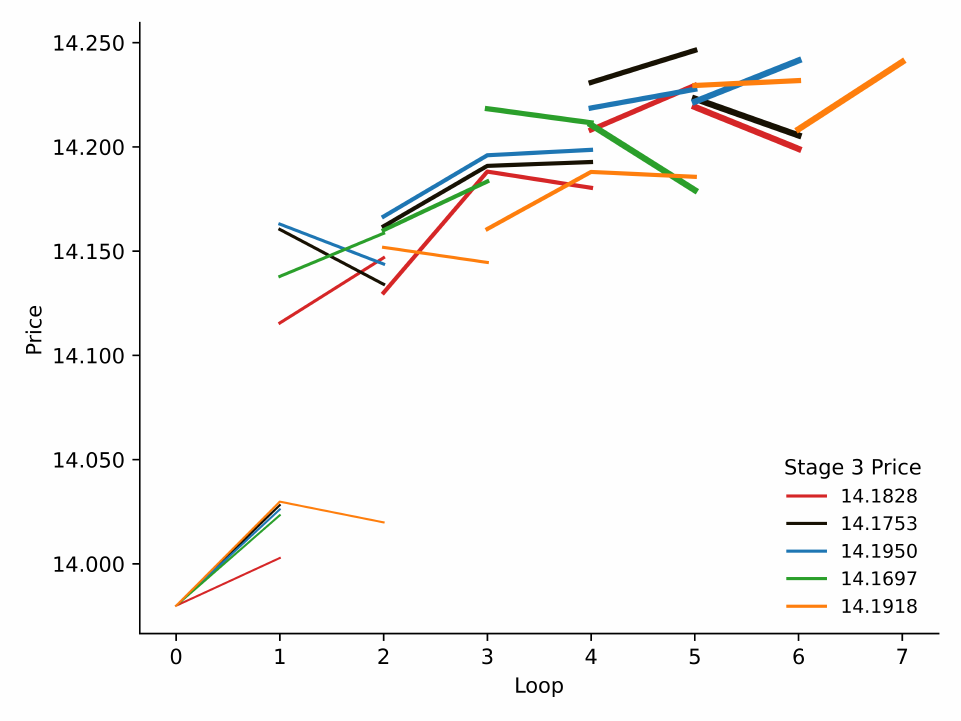}
    \caption{
        Pricing the \texttt{M2} contract from Table~\ref{tab:contracts-tab} under the baseline CARLOS configuration in Table~\ref{tab:Parameter-Setting}. 
        \emph{Left panel (a):} Stopping boundaries (obtained using a marching-squares algorithm from the ADNN $R^{[\ell]}$) $B^{[\ell]}$ at time $t=1$,  after each RL iteration $\ell=0,\ldots, L$ as the step size is refined from $\Delta t^{tr} = \frac{1}{6}$ to $\Delta t^{tr} = \frac{1}{96}$. 
        The final Stage~2 boundary at loop $L = 7$ is drawn with a thicker line. Distinct colors indicate different grid levels $b$, with progressively lighter shades marking repeated loops on the same grid $\bT^{(tr,b)}$. The dashed lines encompass the out-of-the-money region.
        The heatmap (computed via an antialised interpolant) corresponds to the final estimated timing values $\hat\cT^{[L]}(t,x)$. 
        \emph{Right panel (b):} Average validation rewards $\upsilon^{[\ell]}$ across loops $\ell=1,2,\ldots,L$, over five independent RL runs.
        Thicker lines denote finer solver grids $\bT^{(tr,b)}$.
    }
    \label{fig:M2-Contour-Stability}
\end{figure}

As a further illustration, Figure~\ref{fig:M2-Contour-Stability} shows CARLOS solving the 2-d Max Call \texttt{M2} contract from Table~\ref{tab:contracts-tab}. The left panel of Figure~\ref{fig:M2-Contour-Stability}  tracks the changes in the stopping boundary at time $t=1$ over successive learning iterations. Once again we note the expansion of the continuation region as $\Delta t^{ex}$ is decreased. 
The resulting ADNN boundaries resemble those reported by \cite{ValentinTissot, GuanglianLi}.  The right panel illustrates the gradual improvements during Stage~2, tracking the average rewards across the RL loops (cf.~the earlier Figure~\ref{fig:B1}) across five independent runs of CARLOS. Note that due to the adaptive stopping criterion, the resulting number of RL loops $L$ varies across runs, being as low as 5 and as high as 7.  Running our algorithm 5 times, we obtain final prices in the range $[14.170, 14.195]$, just shy of the explicit finite-difference PDE solver value of 14.214 in Table~\ref{tab:PDE-tab-app}. For comparison purposes,  the Bermudan version with $\Delta t = 1/3$ has reference prices of 13.901 and 13.898 as reported in \cite{Becker} and \cite{ValentinTissot}, while CARLOS yields
14.173 (see Table~\ref{tab:Parameter-Setting}), capturing more than 90\% of the materially large gap 
(30 cents or more than 2\%) between the American and Bermudan formulations.

\subsection{Benchmarked Results}\label{sec:Results}

We proceed to evaluate our RL algorithm on a test suite of contracts with a range of payoffs and number of underlying assets. Unlike the vast majority of extant literature, our goal is not to value a Bermudan option with a pre-specified exercise frequency, but the true American contract where exercise can occur at any time. 
To provide a comprehensive assessment, we consider a collection of option contracts in dimensions 1-5, listed in Table~\ref{tab:contracts-tab}. We do not tackle truly high-dimensional problems (which require additional adjustments), but rather focus on realistic settings, emphasizing the aforementioned concepts of time-discretization and efficient learning. 

To assess performance, for low-dimensional tests we benchmark against PDE-based solvers, which can handle very small $\Delta t^{ex}$ but are only tractable in dimensions 1-2\footnote{implementations in 3 dimensions exist, but are cumbersome and not done here.}. These provide a gold-standard deterministic comparator.
Next, we compare to the price estimate based on ADNN that is trained on a medium-grained grid $\Delta t^{tr}$ and  evaluated on Monte Carlo paths at a fine exercise frequency $\Delta t^{ex}$. This setup allows to capture a substantial chunk of the American early exercise premium as described in Section~\ref{subsec:LSMC-Errors},  albeit at a high computational cost. Finally, we compare our results against Bermudan solvers reported in the literature, which serve as reference lower bounds for the  American value and quantify the gains from removing the time discretization.

\begin{table}[ht]
\centering
\begin{tabular}{cc|ccccccc }
    \hline
    Case & Payoff & $d$ & $X_0$ & ${\cal K}$ & $T$ & $\vec\delta$ & $r$ & $\vec\sigma$\\
    \hline\hline
    \texttt{B1} & Basket Put & 1 & 36 &  40 & 1 & 0 & 0.05 & 0.2\\
    \texttt{B2} & Basket Put & 2 & $\{40, 40\}$ & 40 & 1 & $\{0, 0\}$ & 0.06 & $\{0.2, 0.2\}$\\
    \hline
    \texttt{M2.A} & Max Call & 2 & $\{100, 100\}$ & 100 & 3 & $\{0.1, 0.1\}$ & 0.05 & $\{0.2, 0.2\}$\\
    \texttt{M2.B} & Max Call & 2 & $\{100, 100\}$ & 100 & 3 & $\{0.05, 0.15\}$ & 0.05 & $\{0.2, 0.2\}$\\
    \hline
    \texttt{M3} & Max Call & 3 & \makecell{$\{90, 90, 90\}$} & 100 & 3 & $\{0.1, 0.1, 0.1\}$ & 0.05 & \makecell{$\{0.2, 0.2, 0.2\}$}\\
    \hline
    \texttt{M5.A} & Max Call & 5 & \makecell{$\{100, 100, 100,$\\ $100, 100\}$} & 100 & 3 & \makecell{$\{0.1, 0.1, 0.1,$\\ $0.1, 0.1\}$} & 0.05 & \makecell{$\{0.2, 0.2,$\\ $0.2, 0.2, 0.2\}$}\\
    \texttt{M5.B} & Max Call & 5 & \makecell{$\{70, 70, 70,$\\ $70, 70\}$} & 100 & 3 & \makecell{$\{0.1, 0.1, 0.1,$\\ $0.1, 0.1\}$} & 0.05 & \makecell{$\{0.08, 0.16,$\\ $0.24, 0.32, 0.4\}$}\\
    \hline
\end{tabular}
\caption{Specifications of the benchmarked option contracts. Basket Put and Max Call payoffs are as in \eqref{eq:Payoff-Basket-Put}.
For each contract, $d$ denotes the number of underlying assets, $X_0$ the $d$-vector initial condition, $\cal K$ the strike price,  $T$ the maturity, $\vec\delta$ the vector of dividend yields, $r$ the risk-free interest rate and $\vec\sigma$ the vector of volatilities.}
\label{tab:contracts-tab} 
\end{table}

The underlying assets follow a $d$-dimensional Black--Scholes framework. Under the risk-neutral measure (still denoted as $\mathbb{P}$), the asset dynamics are given by
\begin{equation}
    \label{eq:BS1}
    X_t^i = X_0^i \exp\left\{\left(r-\delta_i-\tfrac{\sigma_i^2}{2}\right)t + \sigma_i W_t^i\right\}, 
    \quad i=1,2,\ldots,d,
\end{equation}
where $X_0^i>0$ is the initial value, $\delta_i\ge 0$ the dividend yield, and $\sigma_i>0$ the volatility of the $i^{\text{th}}$ asset; $r\in\mathbb{R}$ is the risk-free rate. All the Brownian motions $W^1,\ldots,W^d$ are independent. 
We consider two payoff families: the arithmetic basket put $h_{bskt}$ and the max call $h_{mxcl}$: 
\begin{equation}
\label{eq:Payoff-Basket-Put}
h_{bskt}(X_t)=\Bigl({\cal K}-\frac{1}{d}\sum_{i=1}^d X_t^i\Bigr)_{+}, \qquad 
h_{mxcl}(X_t)=\Bigl(\max_{1\le i\le d} X_t^i-{\cal K}\Bigr)_{+}.
\end{equation}
When $d=1$, these reduce to the standard put and call payoffs, respectively.

To assess the CARLOS Algorithm~\ref{alg:RL-Basic}, we compare the Stage~2 price estimates with the benchmark prices summarized in Table~\ref{tab:contracts-comparators}. For the ADNN comparator, we report the average price over five independent instantiations, each using the contract-specific parameter configuration in Table~\ref{tab:Parameter-Stage-2}, to reduce run-to-run variability. Literature benchmarks are available for all contracts in Table~\ref{tab:contracts-tab} except \texttt{B1}; the closest setup is studied in \cite{Langrene} and differs only in the risk-free rate, with $r=0.06$. Consequently, for the \texttt{B1} option we use a PDE-based comparator and price the \cite{Langrene} variant with $r=0.06$ for a direct cross-check. Our Crank--Nicolson (CN) scheme with $\Delta t^{\mathrm{PDE}}=\frac{1}{192}$ and $\Delta x=0.02$ yields a deterministic price of $4.4846$ at $r=0.06$, whereas \cite{Langrene} reports an upper bound of $4.4893$ for the Bermudan formulation with $\Delta t^{\mathrm{tr}}=\frac{1}{50}$. 

\begin{table}[ht]
\centering
\[\begin{array}{lcrrrrr}
    \hline
    \text{Contract} & \text{Literature} & \text{PDE} & \text{ADNN} & \text{CARLOS} & \text{ADNN Time} & \text{RL Time} \\
    \hline\hline
    \texttt{B1}   & -                         & 4.601  & 4.583 (0.010)  & 4.592 (0.005)  & 246.38 (1.77) &  11.70 (1.66) \\
    \texttt{B2}   & 1.460  \ \cite{mlOSP}    & 1.478  & 1.468 (0.005)  & 1.474 (0.001)  & 140.57 (0.24)  & 9.37 (0.81)  \\
    \texttt{M2.A} & 13.901 \ \cite{Becker}     & 14.214 & 14.133 (0.067) & 14.171 (0.015) & 297.15 (0.39) & 57.00 (5.11)  \\
    \texttt{M2.B} & 15.575 \cite{ValentinTissot} & 15.777 & 15.615 (0.131)              &  15.711 (0.022)              & 302.93 (0.84) &      51.02 (0.95)         \\
    \texttt{M3}   & 11.278 \ \cite{Becker}     & -      & 11.457 (0.031) & 11.510 (0.008) & 1272.05 (22.97) & 117.91 (4.63) \\
    \texttt{M5.A} & 26.151  \ \cite{ValentinTissot} & -      & 26.401 (0.034)              & 26.55 (0.032)             & 2647.36 (38.86)   &      451.28 (36.26)      \\
    \texttt{M5.B} & 11.810 \ \cite{mlOSP}     & -      & 11.914 (0.073) & 12.009 (0.010) & 780.04 (8.93) & 344.69 (44.41)  \\
        \hline
\end{array}\]
\caption{Benchmark price comparators for the option contracts in Table~\ref{tab:contracts-tab}: PDE-based Monte Carlo price estimates reported in Table~\ref{tab:PDE-tab-app} for contracts in dimension $d \le 2$, ADNN comparators, and the highest report literature sources. Prices are computed using $1.6\times10^6$ Monte Carlo paths at exercise frequency $\Delta t^{ex} = \frac{1}{192}$. ADNN and CARLOS values are averaged over five independent runs; standard deviations are shown in parentheses. 
Parameter configurations are reported in Tables~\ref{tab:Parameter-Stage-2} and~\ref{tab:Parameter-Setting}, respectively. The training times are reported in seconds. }
\label{tab:contracts-comparators}
\end{table}

Besides the benchmark prices in Table~\ref{tab:contracts-comparators}, several contracts are also considered elsewhere.  The \texttt{M2.A} contract has reported prices of $13.897$ in \cite{GuanglianLi} and $13.898$ in \cite{ValentinTissot}. For \texttt{M3}, \cite{mlOSP} reports a price of $11.15$. For \texttt{M5.B}, reported prices include $25.84$ in \cite{mlOSP}, $26.0553$ in \cite{GuanglianLi}, and $26.147$ in \cite{Becker}.

Table \ref{tab:contracts-comparators} presents the results for the 7 contracts.
By using a large $N$ (equivalently, a small training time step $\Delta t^{tr, 0}$),
the ADNN comparators can approach the American-style price,  but this requires a very large number of paths $K$, otherwise the regression predictions are not stable. This refinement comes at a substantial computational cost, especially in dimension $d>2$. Our method not only substantially improves on ADNN but also does so in much shorter runtime. This happens thanks to most of the training done on coarser, hence cheaper grids. Table \ref{tab:contracts-comparators} shows speeds-up of 3-5 times, and sometimes up to 10x, see the M3 contract. 

\section{Algorithm Details and Ablation Studies}\label{sec:RL-details}

In this section we provide additional details of our algorithm and present a set of experiments that test the role of the different tuning parameters. A comprehensive analysis can be found in \cite{Cosmin}. 

All parameter tuning is performed on the 2-dimensional Max Call \texttt{M2} option from Table~\ref{tab:contracts-tab}.  Starting with the baseline configuration  in Table~\ref{tab:Parameter-Setting}, we vary one parameter at a time, reporting post-RL prices, the number of RL loops, and the total training time in seconds. 
Specifically, we tune hyperparameters controlling the training-set size per loop, input selection, time-grid refinement, and the extent of exploration in computing timing values. We also analyze failure modes to illustrate how ill-tuned settings impede learning and lead to low rewards.
 
In all Tables, reported values are averages over five independent runs with different random seeds, with standard deviations given in parentheses.
Final prices are computed using $1.6\times10^6$ Monte Carlo paths at exercise frequency $\Delta t^{ex} = \frac{1}{192}$; the standard error of these estimates is approximately $0.013$.

All numerical computations were performed on the CPU of a single Linux server equipped with an AMD Ryzen Threadripper PRO 5965WX 
and 251 GiB of system memory. The system provides 48 logical CPUs (2 threads per core) on a single NUMA node at up to 4.57 GHz. 

\subsection{Neural Network Architecture}\label{sec:NN-Architecture}
The ADNN in CARLOS is implemented as a fully-connected feedforward neural network $R^\Theta : \mathbb{R}^{d+1} \to \mathbb{R}$ where the additional dimension accounts for the time input. In Stage~1a, we have a similar construction where the input is just the $d$-dimensional state,  $R^{\theta_n} : \mathbb{R}^d \to \mathbb{R}$. The architecture of the network with $I$ layers that consist of $q_i, i=1,\ldots, I-1$ nodes respectively, is defined in standard form \cite{Becker} as:
\begin{equation}
    \label{eq:ADNN-Architecture} 
	R^{\theta} = \varphi_{q_I} \circ a_I^{\theta} \circ \varphi_{q_{I-1}} \circ a_{I-1}^{\theta} \circ \cdots \circ \varphi_{q_1} \circ a_1^{\theta}
\end{equation}
where
$q_I = 1$ is the output dimension, $q_0\in \{d, d+1\}$ is the input dimension and
 $a_i^{\theta} : \mathbb{R}^{q_{i-1}} \to \mathbb{R}^{q_i}, i=1, \ldots,I$ are affine functions with weight matrices $\bA_i \in \mathbb{R}^{q_i \times q_{i-1}}$ and bias vectors $b_i \in \mathbb{R}^{q_i}$:
    $$a_i^{\theta}(x) = \bA_i x + b_i, \quad i = 1, \ldots, I.$$
The parameters $\Theta$ of a network $R^{\Theta}$ comprise the entries of the weight matrices $\bA_1, \bA_2, \ldots, \bA_I$ and the bias vectors $b_1, b_2, \ldots, b_I$. 
Consequently, the dimension of $\Theta$ is given by
\begin{equation}
    \label{eq:NND3} 
	|\Theta| = 1 + q_1 + \ldots + q_{I-1} + q_0 q_1 + \ldots + q_{I-2}q_{I-1} + q_{I-1}. 
\end{equation}

By default we use 3 hidden layers ($I=4$) with equal number $q_i \equiv q$ of hidden nodes, for a total of $|\Theta|=1+(4+q_0)q+2q^2$ of hyperparameters. The number of neurons $q_i$ must be large enough to allow the ADNN to fit the timing value hypersurface. If the widths $q_i$'s are too low, performance suffers; adding nodes beyond a certain level does not translate into materially higher rewards, as fit quality saturates. Theoretically, \cite{gonon2024deep} shows that the number of parameters in a ReLU deep NN surrogate for LSMC ought to scale polynomially in dimension $d$ to maintain a given expressivity error.
We do observe that in higher-dimensional settings wider networks are beneficial and in turn require more training epochs to train. Again, insufficient number of epochs lowers accuracy, but a plateau is quickly reached as epochs are added. Algorithm runtime scales linearly in the number of epochs, so it can be worthwhile to tune that parameter.
    
A variety of choices have been proposed for  the component-wise activation functions
$\varphi_q : \mathbb{R}^q \to \mathbb{R}^q$, including ReLU \cite{Cheridito}, Leaky ReLU \cite{LapeyreBernard,HerreraCalypso}, Tanh \cite{Becker,HerreraCalypso}; we also mention ELU and Swish. On the one hand, the choice of the activation function tends to have a second-order effect on the final price. On the other hand, different activation functions seem to do better for different contracts and moreover $\phi_q$ affects the smoothness of the stopping boundaries: a non-smooth activation like ReLU leads to zig-zaggy, piecewise linear stopping boundaries.   We recommend the Swish function
  \begin{equation}
    \label{eq:Swish} 
    \varphi_q(x_1, \ldots, x_q) := (\frac{x_1}{1+e^{-x_1}} , \ldots, \frac{x_q}{1+e^{-x_q}}),
    \end{equation}    
    or ReLU, $\varphi_q(x_1, \ldots, x_q) := (x_1^{+} , \ldots, x_q^{+}), \ x^+ := \max\{x,0\}$.
The output layer uses the linear activation function $\varphi_{q_I}(x) = x$.

We initialize weights and biases using PyTorch’s built-in uniform initialization, and train using the Adam optimizer based on the mean squared error (MSE) loss function. We take $\eta^{[0]} = 10^{-4}$ as the default initial learning rate. Adam minibatch size has little impact and based on experiments in \cite{Cosmin}, we use batch size of 64 throughout.

\subsection{Selecting Inputs}\label{subsec:RL-Inputs}

Since the primary task of the surrogate is to determine the optimal stopping rule, we wish to preferentially train on inputs that are near the estimated stopping boundary. This principle of ``training in regions where it matters''---that is, focusing on areas that contribute to achieving superior rewards instead of sampling inputs uniformly corresponds to the \emph{exploitation} aspect of our training approach. Cognizant of the respective \emph{exploration} aspect, we also train on inputs beyond the boundary, lest the surrogate makes incorrect stopping decisions in other regions. 
The balancing of the exploitation-exploration tradeoff is essential because neural networks are prone to overexploit and find hyperparameters that exclusively optimize performance in the training region. A further challenge is learning timing values near maturity $T$ where the stopping boundary moves rapidly (see Figure \ref{fig:BP1-Boundaries}), and which calls for additional training inputs in that region.

 With the above considerations in mind, we develop an adaptive multi-pronged sampling strategy that is formalized through a sampling density $p^{[\ell]}(\cdot, \cdot)$ on $\bT^{(tr, b)} \times \cX$. 
To construct $p^{[\ell]}$ that governs the selection of the inputs $(\mathbf{t}, \mathbf{x})^{[\ell]}$ in line 3 of Algorithm \ref{alg:RL-Basic}, we employ a weighted mixture of four components: 
\begin{equation}
    \label{eq:KDE} 
     p^{[\ell]} := \lambda_{+}^{[b]} \cdot p^{\{b, +\}} + \lambda_{-}^{[b]} \cdot p^{\{b, -\}} + \lambda_{\text{exl}}^{[b]} \cdot p^{\{b, \text{exl}\}} + \lambda_{\text{ter}}^{[b]} \cdot p^{\{b, \text{ter}\}}    \qquad  \lambda^{[b]}_{+} + \lambda^{[b]}_{-}+ \lambda^{[b]}_{ \text{exl}} + \lambda^{[b]}_{\text{ter}} =1.  
\end{equation}
In \eqref{eq:KDE}, we balance the exploratory sampling via $p^{\{b, \text{exl}\}}$ with the exploitative boundary-oriented sampling via $p^{\{b, +\}}$ (samples with positive timing values near the estimated stopping boundary) and $p^{\{b, -\}}$ (samples with negative timing values close to the boundary). Moreover, we also explicitly incorporate terminal samples at $t=T$ governed by $p^{\{b, \text{ter}\}}$. The weights $\lambda_{+}^{[b]}$, $\lambda_{-}^{[b]}$, $\lambda_{\text{exl}}^{[b]}$, and $\lambda_{\text{ter}}^{[b]}$ are adaptively adjusted as a function of the grid counter~$b$.

Practically, we implement \eqref{eq:KDE} as kernel-smoothed densities based on \emph{anchor sites}. To construct the anchor sites we generate $P$ i.i.d.~pilot paths $\{\check{\mathbf{x}}\}^{1:P}$, started from the randomized initial condition
\begin{equation}
    \label{eq:Init-Loc} 
    \check{x}_{0}^{p,i} = 
    [{\cal K} 1_{h(X_0) = 0} + X_0 1_{h(X_0)>0} ]\bigl(1 + \beta^i \sigma_i \eps_0^i\bigr), \qquad i = 1,  \ldots, d, \qquad \eps_0^i \sim {\cal N}(0,1).   
\end{equation}
hey start from at-the-money. The randomized initial condition ensures more exploration, namely to provide information about the timing values for $t$ small. 
We furthermore discard any $\check{x}^p_{0}$ that are out of the money, $h( \check{x}^p_{0}) = 0$, or have a non-positive initial timing value $\hat{\T}^{[\ell]}(0, \check{x}^p_{0}) \le 0$ (deep in-the-money). 

To construct the exploitative anchor sites that target the neighborhood of the stopping boundary, we look for  
 time steps $\check{t}^p$ along each pilot path $\check{\mathbf{x}}^p$, where the pathwise action shifts from \textit{continue} to \textit{stop}, i.e., where the current decision is $\phi_{\check{t}^p}^p = 1$ while the previous decision was $\phi_{\check{t}^p-1}^p = 0$.
By construction, timing values are negative at  $(\check{t}^p, \check{x}_{t_n}^p)$---which we use for the negative anchor sites $(\check{\mathbf{t}}, \check{\mathbf{x}})^{\{b, -\}}$---and positive at $(\check{t}_{n-1}, \check{x}_{t_{n-1}}^p)$---used for  $(\check{\mathbf{t}}, \check{\mathbf{x}})^{\{b, +\}}$. Pilot paths that are in-the-money are used for the  exploratory anchor set $(\check{\mathbf{t}}, \check{\mathbf{x}})^{\{b, \text{exl}\}}$. 

Finally, the terminal anchor sites are based on the terminal values of the in-the-money pilot paths $({{T}}, \check{\mathbf{x}})^{\{b, \text{ter}\}}$. To fence against the upward pull in the learned timing values near $T$, we assign them a negative terminal timing value~$y^{b}_{\text{ter}}$ linked to the loss from delayed exercise:
\begin{equation}
    \label{eq:Terminal-TV} 
    y^{b}_{\text{ter}}(T, x^p_T) := c_{ter,0} \cdot c_{ter,1}^b  \cdot  
    \Delta t^{tr, b}. 
\end{equation}
Here, 
$c_{\text{ter},0}, c_{\text{ter},1} $ are tuning parameters. The pseudocode for constructing anchor sets is provided in Algorithm [\ref{alg:RL-Inputs}].  

Given the set of anchor sites, actual training inputs  $(\mathbf{t}, \mathbf{x}),$ are obtained through jittering (kernel-smoothing) the $(\check{\mathbf{t}}, \check{\mathbf{x})}$
with Gaussian noise of $0.01 \sigma_i x_i$ in space and $0.5\Delta t^{tr,b}$ in time. 
To reduce the sample noise of the pathwise payoffs $y^b(t, x_t)$ in \eqref{eq:RL-TV}, we generate $\check{P}=4$ 
paths emanating from each input and use the respective average for ADNN training.

\begin{algorithm}[ht]
\caption{Anchor Set Construction}\label{alg:RL-Inputs} 
\begin{algorithmic}[1]
\Require Regression surrogate $R^{\Theta}$, \# of pilot paths $P$, Solver grid $\bT^{(tr, b)}$, Payoff function $h$.
\State Generate $P$ pilot paths $\{\check{\mathbf{x}}\}^{1:P}$ on $\bT^{(tr, b)}$. Initialize all anchor components to be empty.
\State For all $n=1,\ldots,N, p=1,\ldots, p$ predict timing values $\hat{\T}_n^p \leftarrow \hat{\T}(t_n, \check{x}_{t_n}^p)$ using $R^{\Theta}$, compute immediate payoffs $h_{n}^p \leftarrow h( \check{x}_{t_n}^p)$ and evaluate 
 stopping decisions $\phi_{n}^p \gets \mathbf{1}_{\{\hat{\T}_n^p \le 0\} \cap \{h_{n}^p > 0\}}$ 
\For{$t_n \in \bT^{(tr, b)}$}
    \State Append $(\check{\mathbf{t}}, \check{\mathbf{x}})^{\{b, -\}}$ by $\{(t_n, \check{x}_{t_n}^p) \in \{\check{\mathbf{x}}\}^{1:P}: \phi_{{n-1}}^p = 0, \phi_{n}^p = 1, h_{n}^p > 0 \}$
    \State Append $(\check{\mathbf{t}}, \check{\mathbf{x}})^{\{b, +\}}$ by $\{(t_{n-1}, \check{x}_{t_{n-1}}^p) \in \{\check{\mathbf{x}}\}^{1:P}: \phi_{{n-1}}^p = 0, \phi_{n}^p = 1, h_{n}^p > 0 \}$
    \State Append exploratory sites $(\check{\mathbf{t}}, \check{\mathbf{x}})^{\{b, \text{exl}\}}$ by $\{({t}_n, \check{x}^p_{t_n}) \in \{\check{\mathbf{x}}\}^{1:P}: h^p_n > 0\}$
\EndFor 

\State Save terminal sites $(\check{\mathbf{T}}, \check{\mathbf{x}})^{\{b, \text{ter}\}} \gets \{(T, \check{x}^p_T): h( \check{x}^p_T) > 0\}$
\State\Return{$(\check{\mathbf{t}}, \check{\mathbf{x}})^{\{b, +\}}, (\check{\mathbf{t}}, \check{\mathbf{x}})^{\{b, -\}}, (\check{\mathbf{t}}, \check{\mathbf{x}})^{\{b, \text{exl}\}}, (\check{\mathbf{T}}, \check{\mathbf{x}})^{\{b, \text{ter}\}}$}
\end{algorithmic}
\end{algorithm}

\begin{figure}[!htb]
    \centering
    \subfloat[Inputs on the solver grid with $\Delta t^{tr} = \frac{1}{12}$. 413 exploratory, 269 with positive timing value,
269 with negative timing value, and 50 terminal. \label{fig:BP1-input-loop-1}]{
        \includegraphics[width=0.475\textwidth,trim=0.25in 0in 0.25in 0.5in]{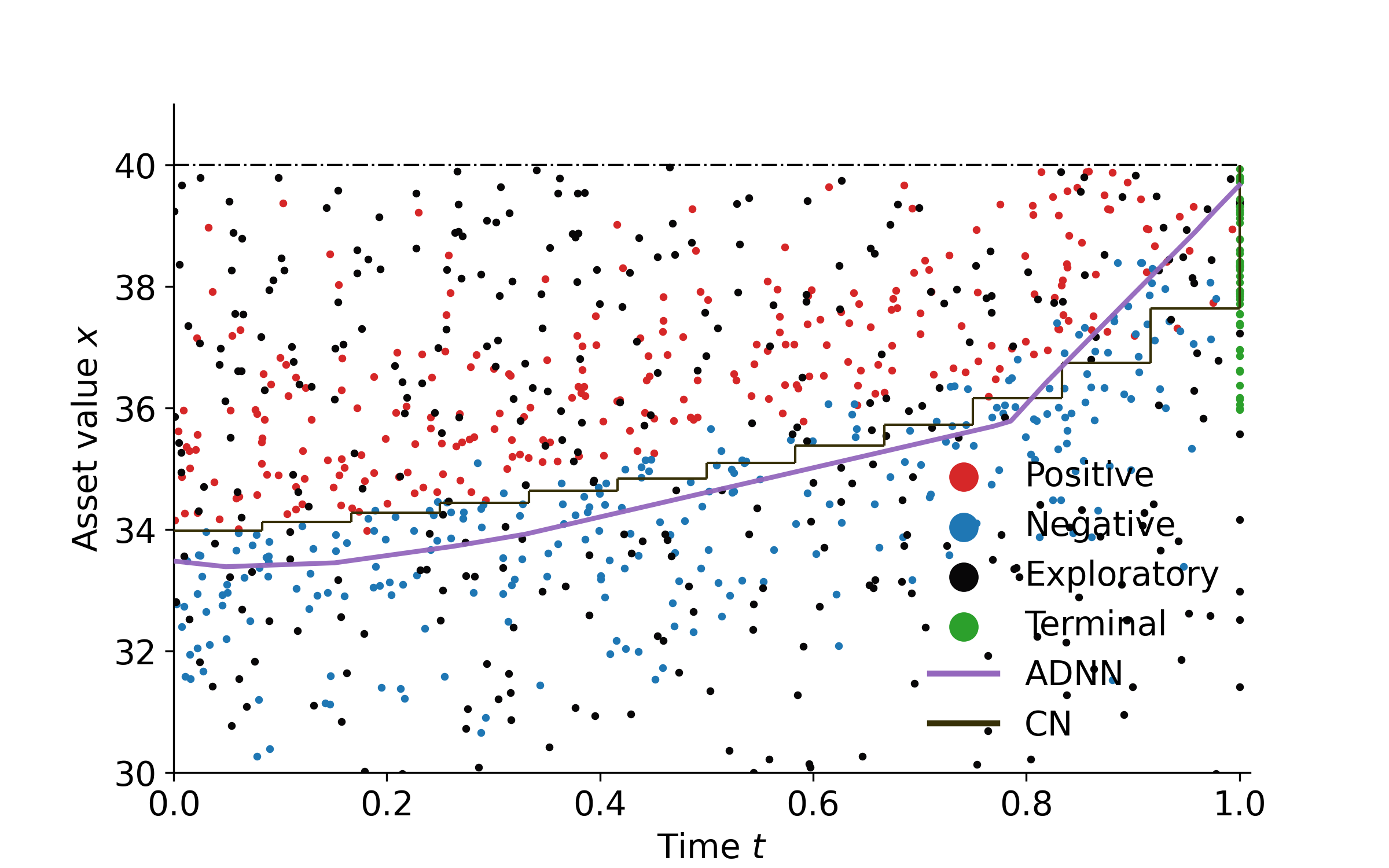}}
    \hfill
    \subfloat[Inputs on the solver grid with $\Delta t^{tr} = \frac{1}{96}$. 174 exploratory, 388 with positive timing value,
388 with negative timing value, and 50 terminal. \label{fig:BP1-input-loop-31}]{
        \includegraphics[width=0.475\textwidth,trim=0.25in 0in 0.25in 0.5in]{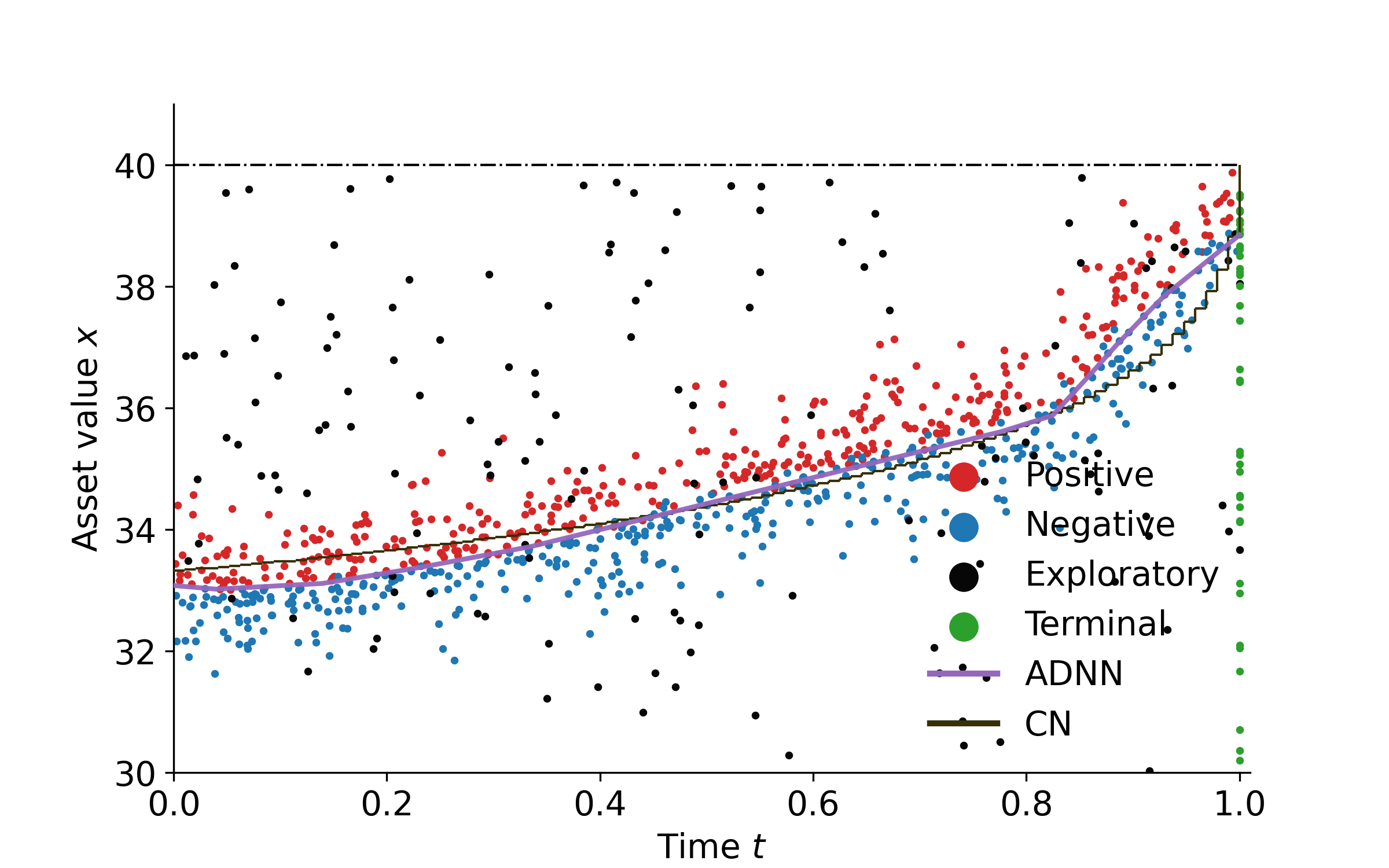}
    }

    \caption{Input sets $(\mathbf{t}, \mathbf{x})^{[\ell]}$  for the \texttt{B1} contract  using the parameter configuration in Table~\ref{tab:Parameter-Setting}. There are $1{,}000$ distinct training inputs categorized into: {exploratory} $(\check{\mathbf{t}}, \check{\mathbf{x}})^{\{b, exl\}}$, {positive} timing value $(\check{\mathbf{t}}, \check{\mathbf{x}})^{\{b, +\}}$, {negative} timing value $(\check{\mathbf{t}}, \check{\mathbf{x}})^{\{b, -\}}$, and {terminal} $(\check{\mathbf{t}}, \check{\mathbf{x}})^{\{b, ter\}}$ sampling mixture components. }
    \label{fig:BP1-inputs-shift}
\end{figure}

Figure \ref{fig:BP1-inputs-shift} visualizes the training inputs $(\mathbf{t}, \mathbf{x})$ across learning loops for the 1-dimensional Bermudan Put option \texttt{B1} from Table~\ref{tab:contracts-tab}. In the left panel, $\Delta t^{tr} = 1/12$ and the training set is more spread out; in the right panel we work with the finest grid with $\Delta t^{tr}=1/96$.
As the time grid is refined, the exploitation inputs---with either positive or negative timing values---concentrate around the estimated stopping boundary $B^{[\ell]}$ because $\check{x}_{t_n}^p$ gets closer to $\check{x}_{t_n-1}^p$ in lines 4 and 5 of Algorithm \ref{alg:RL-Inputs}.

\begin{remark}
The above discussion emphasized the benefit of spatially targeting the region around the stopping boundary. For the temporal aspect in terms of $t$, there is also a non-uniform target, namely the distribution of the stopping time $\tau^*$. Figure \ref{fig:tau-distribution} in the Appendix shows the density of $\tau^* \in [0,T]$ (which heavily depends on the initial starting location) for the B1 contract. For training purposes, we seek to make sure that $\check{\mathbf{t}}$ has a strictly positive  density on the entire $[0,T]$. Observe that if $X_0$ is out-of-the-money then the probability that $\tau^*$ is close to zero is negligible, conflicting with the above aim. This is one of the reasons to randomize the starting $x^p_0$ of the pilot paths, see the resulting distribution of $\check{\mathbf{t}}$ in the right panel of Figure \ref{fig:tau-distribution}. A further target could be either to take $\check{\mathbf{t}}$ roughly uniform on $[0,T]$ (our default choice) or to preferentially sample closer to $T$ when stopping decisions are most frequently made in many contracts like the max Calls where time is needed for the paths to reach the upper-right quadrant that has the largest impact on option value.
\end{remark}

\textbf{Exploratory Coverage:} In Algorithm~\ref{alg:RL-Inputs}, anchor sites at $t_n$ are based on the pilot paths and reflect the underlying distribution of $X_{t_n} | X_0$.
This can be a limiting factor for exploration; in our examples the distribution of $X_t$ is log-normal which has a  long right tail. To be concrete, the $2{,}500$ distinct exploratory inputs for the \texttt{M2.A} contract span  $[33.55, 288.36] \times [32.49, 244.66]$. By contrast, the $1.6 \times 10^6$ testing paths span a much broader range, $[14.74, 423.22] \times [15.62, 473.35]$. This mismatch between training and testing set implies that the ADNN stopping rule is inaccurate for very high $X_1,X_2$ values, e.g., along the diagonal $X_1,X_2 \in [200,400]^2 $, which has a disproportionally high impact on option value as this is precisely the  high-payoff region.

To address this, it may be beneficial to modify the sampling scheme for exploratory training inputs.
For instance, we consider a mixture of the underlying log-normal density and a uniform sampling near the
diagonal---to obtain more informative training inputs that can expand the learned boundary and sharpen its geometry. Accordingly, we let 
\begin{equation}
\label{eq:Exploratory-Hybrid}   
p_{\text{hybrid}}^{\{b,\text{exl}\}} := (1-c_{\text{band}})\,p^{\{b,\text{exl}\}} + c_{\text{band}}\,\mathrm{Unif}([1,3]\times \mathcal{X}_{\text{band}}),
\end{equation}
where $c_{\text{band}}$ controls the weight of the contract-specific exploration component and $c_{\text{width}}$ controls the width of the diagonal band
\begin{equation}
\label{eq:Exploratory-Tube}   
\mathcal{X}_{\text{band}} = \left\{(x_1, x_2) \in \mathcal{X}: |x_1- x_2| \le c_{\text{width}},\; x_1, x_2 \in [x_{\min}, x_{\max}]\right\}.
\end{equation}
 A simple manual calibration is to set~$x_{\min}=\cK$ to match the contract's strike price and choose $x_{\max}=300, c_{\text{width}}=30$. To stay on the conservative side, a uniform distribution over the time window 
$[1,3]$ is used as reaching the diagonal band is unlikely early on.

\textbf{Tuning Training Set Geometry:}  The exploration-exploitation balance is controlled by the weights $\lambda^{[b]}_{+}$ and $\lambda^{[b]}_{-}$ in Equation~(\ref{eq:KDE}) that define the sampling density $p^{[\ell]}$. 
As the solver-grid step decreases, the anchor inputs in Algorithm~[\ref{alg:RL-Inputs}] are drawn closer to the stopping boundary, so exploitation is intrinsically amplified as $\Delta t^{tr}$ shrinks. While advantageous for improving the ADNN fit around the boundary, over-exploitation limits the ability of ADNN to self-correct and can make the RL prone to error accumulation. Moreover, at the earlier RL iterations the boundary may shift substantially, especially during grid transitions, and selecting inputs based on the previous solver grid can then misdirect learning. At the same time, exploration on fine grids is also inefficient, since the stopping boundary stabilizes once the grid step size $\Delta t^{\mathrm{tr}}$ becomes small.

These considerations motivate a \emph{gradual} reallocation of training effort from exploration to exploitation. 
Let $c_{\text{expl}}$ denote the proportion of exploratory inputs that is reallocated to exploitation after each solver-grid transition. When advancing the grid index from $b$ to $b + 1$, we update the weights used to construct~$p^{[\ell]}$ in Equation~\eqref{eq:KDE} as
\begin{align}\label{eq:lambda_update}
\lambda_{\text{exl}}^{[b+1]} := (1 - c_{\text{expl}})\lambda_{\text{exl}}^{[b]}, \quad 
\lambda_{+}^{[b+1]} := \lambda_{+}^{[b]} + \frac{c_{\text{expl}}}{2}\lambda_{\text{exl}}^{[b]}, \quad
\lambda_{-}^{[b+1]} := \lambda_{-}^{[b]} + \frac{c_{\text{expl}}}{2}\lambda_{\text{exl}}^{[b]},
\end{align}
while keeping $\lambda_{\text{ter}}^{[b]}$ fixed. 
Setting $c_{\text{expl}}=0$ recovers the baseline setting in which the weights remain constant throughout the RL loops.

\begin{table}[ht]
\centering
\begin{tabular}{l|rrrr}
\hline
$(\lambda_{\text{exl}}^{[0]}, \lambda_{+}^{[0]}, \lambda_{-}^{[0]}, \lambda_{\text{ter}}^{[0]})$ & $c_{expl}$ &  Price & Loops $L$ & Runtime \\
\hline\hline
  $(0.95, 0, 0, 0.05)$ & 0 & $14.1438_{(0.014)}$ & $11.2_{(1.17)}$ & $64.20_{(4.76)}$ \\
   $(0.55, 0.2, 0.2, 0.05)$ & 0 &  $14.1641_{(0.013)}$ & $8.6_{(1.62)}$ & $55.36_{(3.86)}$ \\
  $(0.35, 0.3, 0.3, 0.05)$ & 0 & $14.1775_{(0.007)}$ & $10.6_{(3.07)}$ & $61.05_{(10.32)}$ \\
 $(0.15, 0.4, 0.4, 0.05)$ & 0 & $14.1748_{(0.007)}$ & $8.2_{(2.04)}$ & $56.13_{(7.09)}$ \\
  $(0, 0.475, 0.475, 0.05)$ & 0 & $14.1694_{(0.015)}$ & $8.0_{(1.67)}$ & $53.65_{(4.48)}$ \\ \hline
   $(0.5, 0.2, 0.2, 0.1)$ & 0 &  $14.1486_{(0.015)}$ & $8.8_{(2.32)}$ & $54.60_{(5.90)}$ \\
    $(0.6, 0.2, 0.2, 0.0)$ & 0 & $14.1526_{(0.013)}$ & $10.8_{(2.23)}$ & $60.44_{(10.42)}$ \\
  $(0.35, 0.2, 0.4, 0.05)$ & 0 & $14.1725_{(0.005)}$ & $7.8_{(1.94)}$ & $53.82_{(6.51)}$ \\ 
\hline
 $(0.55, 0.2, 0.2, 0.05)$ & 0.25 & $14.1777_{(0.009)}$ & $9.6_{(2.33)}$ & $59.16_{(8.11)}$ \\
  $(0.55, 0.2, 0.2, 0.05)$ & 0.5 &  $14.1698_{(0.018)}$ & $9.6_{(3.38)}$ & $59.16_{(9.47)}$ \\
 $(0.35, 0.3, 0.3, 0.05)$ & 0.25 & $14.1764_{(0.007)}$ & $8.6_{(2.24)}$ & $54.51_{(5.51)}$ \\
$(0.35, 0.3, 0.3, 0.05)$ &  0.5 & $14.1762_{(0.006)}$ & $10.0_{(1.79)}$ & $59.43_{(6.00)}$ \\
\hline
\end{tabular}
\caption{Impact of the sampling  weights $(\lambda_{\text{exl}}$, $\lambda_{+}$, $\lambda_{-}, \lambda_{\text{ter}})$ in \eqref{eq:KDE} and the reallocation factor $c_{\text{expl}}$ for balancing exploration and exploitation, for the \texttt{M2} Max Call contract in Table~\ref{tab:contracts-tab}. }
\label{tab:Stage3-Input-Selection}
\end{table}

Table~\ref{tab:Stage3-Input-Selection} shows the impact of varying the weights $\lambda_{\text{exl}}$, $\lambda_{+}$, $\lambda_{-}, \lambda_{\text{ter}}$ and the tuning parameter $c_{\text{expl}}$ for the \texttt{M2.A} contract, yielding several take-aways.
First, terminal inputs are helpful: $\lambda_{\text{ter}}^{[0]}=0$ does substantially worse than $\lambda_{\text{ter}}^{[0]}=0.05$ ($\lambda_{\text{ter}}^{[0]}=0.10$ is comparable but leaves fewer interior samples). 
Second, the two extremes of no-exploitation $\lambda_{+}^{[0]}=\lambda_{-}^{[0]}=0$ or no-exploration $\lambda_{\text{exl}}^{[b]}=0$ perform relatively poorly. Eliminating exploitation under-samples the region where ADNN accuracy matters most and makes RL less efficient, yielding the lowest price. Conversely, a fully exploitative training with $\lambda_{\text{exl}} = 0$ confines learning to the boundary region. This inhibits self-correction, providing insufficient information to adjust timing values away from the obsolete boundary, leading to lower rewards. 
Third, we find no benefit from asymmetrically sampling within the stopping/continuation regions, $\lambda_{+}^{[0]}<\lambda_{-}^{[0]}$. 
Fourth, a gradual re-allocation towards exploitation $c_{\text{expl}}>0$ is beneficial and moreover makes the performance less sensitive to precise $\lambda_{\text{exl}}^{[0]},\lambda_{+}^{[0]},\lambda_{-}^{[0]}$ values.

In conclusion, we recommend 
a balanced allocation across exploration, exploitation (split evenly between $\lambda_{+}$ and $\lambda_{-}$), and terminal fencing, setting $\lambda_{\text{exl}}^{[0]}=0.55$, $\lambda_{+}^{[0]}=\lambda_{-}^{[0]}=0.2$, and $\lambda_{\text{ter}}^{[0]}=0.05$, and choosing $c_{\text{expl}}=0.25$ to gradually reallocate exploratory mass toward exploitation as the solver grid is refine.

\subsection{Exploratory Stopping}\label{subsec:RL-Outputs}
 
Accurately evaluating timing values becomes challenging as we transition to progressively finer time grids $\bT^{(ex, b)}$. In particular, near the stopping boundary $B^{[\ell]}$  timing values may switch signs, cf.~the shifting stopping boundaries in Figure~\ref{fig:BP1-Boundaries}. 
Recall the stopping decision map $\phi^{[\ell]}$ from \eqref{eq:Stop-Decision} (now based on the ADNN $\hat{\T}^{[\ell]}$) and consider  a generic path $\{x_t\}$ starting from $(t_{\text{init}}, x_{t_{\text{init}}})$. If  $(t_{\text{init}}, x_{t_{\text{init}}})$ falls within the stopping region $\cS^{[\ell]}$, the default would be to stop immediately, leading to $\tau = t_{\text{init}}$. 
This implies that any input with $\hat{\T}^{[\ell]}(t, x) < 0$ will be assigned a negative training output $y$, and hence will (almost surely) end up in the stopping region $\cS^{[\ell+1]}$ of the next-iteration ADNN $\hat{\T}^{[\ell+1]}$. As a result, the naive implementation would cause the continuation region to be non-expanding across the RL rounds. 

To mitigate this, we introduce ``delayed'' stopping, with the goal of adjusting the timing values of training inputs in the ``old'' stopping region. Such exploratory stopping allows a path originating in $\cS^{[\ell]}$ to continue for a bit, in the hopes of crossing into the continuation region which then enables bonified stopping.  Let $\xi^b$ be the first time a path $\{x_t\}$, starting from $(t_{\text{init}}, x_{t_{\text{init}}})$ and progressing along the exercise grid $\bT^{(ex, b)}$, enters $\cC^{[\ell]}$:
\begin{equation}
    \label{eq:First-Time} 
    \xi^b := \min \left\{t \in \{ t_{\text{init}}\} \cup \bT^{(ex, b)} : \phi^{[\ell]}(t, x_t) = 0, t \ge t_{\text{init}} \right\} \wedge T.
\end{equation}
If the path starts in $\cC^{[\ell]}$, then $\xi^b = t_{\text{init}}$; otherwise, $\xi^b > t_{\text{init}}$ and $\xi^b = T$ for a path that never leaves $\cS^{[\ell]}$.
We then define $\zeta^b$ as the first time the path $\{x_t\}$ transitions from the continuation region into the stopping region
\begin{equation}
    \label{eq:Early-Stop} 
    \zeta^b := \min \left\{t > \xi^b, t \in \bT^{(ex, b)}: \phi^{[\ell]}(t, x_t) = 1 \right\}.
\end{equation}
For trajectories originating within $\cS^{[\ell]}$, delayed stopping means that the path is allowed a waiting period of length $\Delta^b_{\text{wait}}$ to exit into $\cC^{[\ell]}$; otherwise, it is stopped once the delay expires. 
Hence, the path is stopped at the minimum of $\zeta^b$ from \eqref{eq:Early-Stop} and 
\begin{equation}
    \label{eq:Delay-Stop} 
    \mathring{\tau}^b := \min\{t > t_{\text{init}} + \Delta^b_{\text{wait}}: t \in \bT^{(ex, b)}\} \wedge T,
\end{equation}
which is the first time point in $\bT^{(ex,b)}$ after $\Delta^b_{\text{wait}}$. The resulting payoff is set to \begin{equation}
    \label{eq:Outputs-DPF} 
    y_{\text{dpf}}^{b}(t_{\text{init}}, x_{t_{\text{init}}}) := 
    \begin{cases}
        e^{-r(\zeta^{b} - t_{\text{init}})} h( x_{\zeta^b}) & \text{if } \xi^{b} \le \mathring{\tau}^b, \\
        e^{-r(\mathring{\tau}^{b} - t_{\text{init}})} h( x_{\mathring{\tau}^b}) & \text{otherwise}.
    \end{cases}
\end{equation} 
If the path $\{x_t\}$ starts in, or enters $\cC^{[\ell]}$ before $\mathring{\tau}^{b}$, it is stopped at the entrance time $\zeta^{b}$; otherwise, it is stopped at $\mathring{\tau}^{b}$, after the waiting period $\Delta^b_{\text{wait}}$ has elapsed.

Exploratory stopping can be understood via a traffic light analogy, captured by labels $z_t \in \{0,1,2\}$, interpreted as yellow, green, and red signals. A path that starts in $\cC^{[\ell]}$ is labeled green, $z_{t_{\text{init}}} = 1$, while one that originates in $\cC^{[\ell]}$ is initially labeled yellow, $z_{t_{\text{init}}} = 0$.
The label transitions follow three rules, illustrated for the \texttt{B1} contract in Figure \ref{fig:BP1-Paths-Color}.
First, when a yellow path enters the continuation region at hitting time $\xi^b$, it turns green: $z_{\xi^b} = 1$, cf.~the green dot on path~1.
Second, when a green path enters the stopping region, it becomes red: $z_{\zeta^b} = 2$, see the red dot marking  $\zeta^b$ on path~2. Third, when a yellow path—i.e., $z_{t_{\text{init}}} = 0$ remains in $\cS^{[\ell]}$ beyond the waiting period $\Delta^b_{\text{wait}}$, it is set to red: $z_{\mathring{\tau}^b} = 2$ and stopped. This is recorded on path~3, where the gray dashed vertical line indicates the deadline $t+\Delta t^b_{\text{wait}}$  and  $\mathring{\tau}^b$ is shown by a red dot. 
 Algorithm~\ref{alg:Evaluation} outlines how the labels $z_t$ guide the computation of $y_{\text{dpf}}^{b}$ in Equation~(\ref{eq:Outputs-DPF}).

\begin{algorithm}[ht]
\caption{Discounted Delayed Payoff Evaluation}\label{alg:Evaluation} 
\begin{algorithmic}[1]
\Require Regression surrogate $R^{\Theta}$, Path $\{{x}_{t_i}\}$ started at $t_{\text{init}},x_{t_{\text{init}}}$, Exercise grid $\bT^{(ex)}$, Exploration window $\Delta_{\text{wait}}$,
\textit{Path and contract parameters} e.g., Payoff function $h$, Discount rate $r$.  
    \State Set initial label $z\gets 1 - \phi^\Theta(t_{\text{init}}, x_{t_{\text{init}}})$
    \For{$t_n \in \{t \in \bT^{(ex)} : t \ge t_{\text{init}}\}$}
        \State Compute stopping decisions $\phi_n \gets \phi^\Theta(t_n, x_{t_n})$ based on $R^{\Theta}$ by Equation (\ref{eq:Stop-Decision})
        \State Set new label $z_{n} \gets z + \phi_n \cdot \mathbf{1}_{\{z = 1\}} + (1 - \phi_n) \cdot \mathbf{1}_{\{z = 0\}}$
        \State Get $y_{\text{dpf}} \gets e^{-r(t_n - t_{\text{init}})} h(x_{t_n}) \cdot (\mathbf{1}_{\{z_{n}  = 2, z = 1\}} + \mathbf{1}_{\{z_{n}  = 0, t_n - t_{\text{init}}\ge \Delta^b_{\text{wait}}\}})$
        \State Update the label $z \gets z_{n}$
    \EndFor 
\State \Return{$y_{\text{dpf}}$}
\end{algorithmic}
\end{algorithm}

The timing value for each training input $(t, x_t) \sim p^{[\ell]}$ is given by the difference between the discounted delayed payoff of a path $\{x_t\}$ starting at $(t, x_t)$ that progresses along the grid $\bT^{(ex, b)}$ and its immediate payoff:
\begin{equation}
    \label{eq:RL-TV} 
    y^{b}(t , x_t) = y_{\text{dpf}}^{b}(t , x_t) - h( x_t).
\end{equation}

To regulate exploration in $\cS^{[\ell]}$ we shorten $\Delta^b_{\text{wait}}$ when time to maturity $T - t_{\text{init}}$ is small or when a finer solver grid provides more stopping opportunities, setting it according to delayed ratio $c_{dlst}$
\begin{equation}
    \label{eq:Delta-Wait} 
    \Delta_{\text{wait}}^{b}(t) := \Delta_{\text{wait}}(t; \Delta t^{tr, b}) = (c_{\text{dlst}})^{b+1} \Delta t^{tr, b}\left(1 - \frac{t}{T}\right).
\end{equation}

Table \ref{tab:Stage3-RL-Pairs} shows the impact of  $c_{\text{dlst}}$. Suppressing any delayed stopping $c_{\text{dlst}}=0$ leads to premature stopping as the expansion of $\cC^{[\ell]}$ is largely removed. 
Taking very large 
$c_{\text{dlst}} = 2$ is also counter-productive, as excessive $\Delta^b_{\text{wait}}$ 
mislearns the ultimate stopping decision. Both produce poor prices.
To allow sufficient exploration within $\cS^{[\ell]}$ without introducing overly long waiting periods~$\Delta^b_{\text{wait}}$, we select $c_{\text{dlst}} \in[1.1, 1.3]$ as a balanced choice. 
In \cite{Cosmin} we also experimented with nonlinear dependence on time-to-maturity, such as using $(1-t/T)^2$ in \eqref{eq:Delta-Wait}, but this was ultimately rejected.

The exploration window $\Delta^b_{\text{wait}}$ introduces discontinuities in timing values. Supplementary Figure~\ref{fig:BP1-ADNN-Spline-Loop} illustrates this artifact of exploratory stopping for the \texttt{B1} option. At the boundary---identified as the zero-timing-value contour of the ADNN $R^{[0]}$---the timing values jump due to the enforced exploration on yellow-labeled paths. Paths~4 and 5 in Figure~\ref{fig:BP1-Paths-Color} exemplify this: they have the same underlying trajectory, shifted so that their origins lie symmetrically across the boundary. 
The green-labeled path~4 is stopped at $\zeta$, while the yellow-labeled path~5 remains in the stopping region and is stopped at $\mathring{\tau}$. 
In this case, because $\zeta < \mathring{\tau}$, Equation~(\ref{eq:RL-TV}) assigns different timing values $y$ to the two paths; if the stopping times coincided, the values would be identical. The two-piece spline in Figure~\ref{fig:BP1-ADNN-Spline-Loop} is constructed on timing values $y$ from Equation~(\ref{eq:RL-TV} and shows that this creates a substantial gap right around $B^{[\ell]}$.
By construction, the ADNN fits a continuous surrogate, smoothing the above discontinuity and  producing timing values that distort the stopping boundary, shown by the dashed vertical black line in Figure~\ref{fig:BP1-ADNN-Spline-Loop}. The discontinuity dinimishes  as $\Delta^b_{\text{wait}}$ shrinks, which is the motivation to reducing the exploration window as $\bT^{(ex, b)}$ is refined.

\subsection{Stage Allocation}\label{subsec:Stage-Allocation}
The quality of Stage~1 shapes the effectiveness of the RL algorithm in Stage~2. If the initial ADNN $R^{[ 0]}$ is poorly trained, the CARLOS algorithm~[\ref{alg:RL-Basic}] does not converge, so the Stage~1a data has to be sufficiently large, and the initial ADNN training in Stage~1b must be sufficiently comprehensive. Conversely, over-allocating computational resources to Stage~1 leaves little room for further improvement in Stage~2 and incurs long runtimes. 
To strike this balance, two parameters need to be tuned: the number of  paths $K$—which determines the size of the data used to train the ADNN $R^{[ 0]}$, and the initial grid-step size~$\Delta t^{ex,0}$. As discussed in Section \ref{subsec:LSMC-Errors}, errors in the LSMC backpropagate, so reducing $\Delta t^{ex,0}$ does not necessarily yield better data.

As shown in supplementary Table~\ref{tab:Stage1-LSMC}, moderately large values of $K$ improve Stage~1 data quality and yield higher final prices. 
Since the LSMC cost scales linearly with $K$, doubling $K$ say from $2 \cdot 10^4$ to $4 \cdot 10^4$ doubles that portion of the runtime but delivers only minimal improvement on the stopping rule.

The number of input–output pairs $M$ controls per-loop training and 
using too few training inputs may lead to instability due to overfitting the ADNN on sparse data \cite{Mnih}. 
Conversely, using excessively large $M$ increases runtime while yielding only marginal improvements in timing-value estimates and rewards.
For the \texttt{M2} contract, Table~\ref{tab:Stage3-RL-Pairs} indicates that $M = 20{,}000$ strikes this balance, since runtime rises sharply for larger $M$ without a commensurate increase in final prices.

\subsection{Grid Transition}\label{sec:refine-grid}
To guide the learning,  we monitor, assess, and adjust the ADNN in each learning loop using the rewards $\Upsilon^{[\ell]}_v$ over the $v=1,\ldots, V$ validation paths $\{\mathbf{x}\}^{1:V}$ using the grid $\bT^{(ex, b)}$ and the ADNN $R^{[\ell]}$. Achieving full convergence for each solver frequency $\Delta t^{tr, b}$ is unnecessary, as the marginal benefits of further accuracy at an intermediate level do not justify the additional computational cost. To decide whether to transition to the next grid level, we 
check whether zero belongs to the interval
\begin{equation}
    \label{eq:SI} 
    0 \in [\bar D^{[\ell]} - z_{1-\alpha/2}\hat\sigma_{D}^{[\ell]}, \bar D^{[\ell]} + z_{1-\alpha/2}\hat\sigma_{D}^{[\ell]} ],    
\end{equation} 
\noindent where $\bar D^{[\ell]}$ and $\hat\sigma_{D}^{[\ell]}$ are 
the mean and standard deviation of ${D}_v^{[\ell]}:=\Upsilon^{[\ell]}_v - \Upsilon^{[\ell-1]}_v$, $v=1,\ldots,V$. 

A very large fraction (as many as 95\% of the paths) of $D_{v}^{[\ell]}$'s are zero. 
First, many validation paths never yield a positive payoff. Second, some paths receive the same reward $\Upsilon^{[\ell]}_v = \Upsilon^{[\ell-1]}_v$ in consecutive loops because ADNN parameter updates do not alter the stopping decisions in Equation~\eqref{eq:Stop-Decision}. 
Accordingly, we use a Delta method to separately account for the variance in the proportion of zeros and the variance of the non-zero $D_v$'s. 
We exploit the decomposition $D_v = I_{\text{nz}} D_{\text{nz}}$, where $I_{\text{nz}}$ is an indicator of $D_v \neq 0$, and $D_{\text{nz}}$ denotes the reward difference conditional on $D_v \neq 0$. 
Let $p = \mathbb{P}(D_v \neq 0)$, and denote $\mu_{\text{nz}}:=\mathbb{E}[D_{\text{nz}}]$ and $\sigma_{\text{nz}}^2:=\mathrm{Var}(D_{\text{nz}})$. By the law of total expectation,
\begin{align}
    \label{eq:reward-difference-mean} 
    \mathbb{E}[D_v]
    &= \mathbb{E}\!\left[D_v\mid D_v \neq 0\right]\mathbb{P}(D_v \neq 0) 
    + \mathbb{E}\!\left[D_v \mid D_v = 0\right]\mathbb{P}(D_v = 0) 
    = p\,\mu_{\text{nz}}; \\
\label{eq:reward-difference-second-moment}
    \mathbb{E}[D^2_v]
    &= \mathbb{E}\!\left[D^2_v \mid D_v \neq 0\right]\mathbb{P}(D_v \neq 0)
    = p\,\mathbb{E}\!\left[D_{\text{nz}}^2\right]
    = p\left(\sigma_{\text{nz}}^2 + \mu_{\text{nz}}^2\right).
\end{align}
It follows that the variance of the i.i.d.~sample average $\bar{D}^{[\ell]} =\frac{1}{V} \sum_v D_v^{[\ell]}$ is
\begin{equation}
    \label{eq:reward-difference-sample-variance} 
    \mathrm{Var}\left(\bar{D}^{[\ell]}\right) =  \mathbb{E}\big[(\bar{D}^{[\ell]})^2 \big] - \mathbb{E} \big[\bar{D}^{[\ell]} \big]^2 
    = \frac{p}{V}\sigma_{\text{nz}}^2+\frac{p(1-p)}{V}\mu_{\text{nz}}^2.
\end{equation}
Replacing $p$, $\mu_{\text{nz}}$, and $\sigma_{\text{nz}}^2$ in Equation~\eqref{eq:reward-difference-sample-variance} with their sample counterparts $\hat p^{[\ell]}$, $\bar{D}_{\text{nz}}^{[\ell]}$, and $(s^{[\ell]}_{\text{nz}})^2$ gives the estimated standard error $\hat\sigma^{[\ell]}=\sqrt{\widehat{\mathrm{Var}}(\bar{D}^{[\ell]})}$. 
The resulting normal-approximation confidence interval at level $100(1-\alpha)\%$  is
$    \bar{D}^{[\ell]} \pm z_{1-\alpha/2}\hat\sigma^{[\ell]}$,    
where $z_{1-\alpha/2}$ denotes the $(1-\alpha/2)$-quantile of the standard normal distribution. Unless stated otherwise, we take $\alpha=0.05$, resulting in a $95\%$ stopping interval. If the interval in Equation~\eqref{eq:SI} contains zero, the reward change is not significant and the grid is advanced to the next level.

\subsection{Grid Schedule}\label{subsec:RL-Parameters}

As a last ingredient of CARLOS we discuss the construction of the grid schedule $\bT^{(tr,b)},b=0,\ldots, B$. Recall that the coarsest grid $\bT^{(tr,0)}$ is used in Stage~1 to generate good-enough training data for fitting the initial ADNN, while Stage~2 refines the timing value estimates on progressively denser grids. In constructing $\{\bT^{(tr, b)}\}_{b=0}^{B}$, we must both limit concept drift and exploit opportunities to speed up the RL algorithm~\ref{alg:RL-Basic}. 
In particular, we prefer maximal progress on coarse solver grids, where training is computationally inexpensive, while limiting iterations on fine grids, where each loop is costly due to high overhead of generating training samples $\by$. The resulting considerations are investigated in Table~\ref{tab:Stage3-RL-Grid-Schedule}.

\begin{table}[ht]
\centering
\renewcommand{\arraystretch}{1.2}
\begin{tabular}{r|rrr}
\hline
Step Size Schedule $\Delta t^{tr,b}$ & Price & Loops $L$ & Runtime \\
\hline\hline
$(\frac{1}{3}, \frac{1}{6}, \frac{1}{12}, \frac{1}{24}, \frac{1}{48}, \frac{1}{96}, \frac{1}{192})$ & $14.1552_{(0.008)}$ & $11.6_{(2.33)}$ & $68.96_{(11.58)}$ \\
$(\frac{1}{3}, \frac{1}{6}, \frac{1}{12}, \frac{1}{24}, \frac{1}{48}, \frac{1}{96})$ & $14.1616_{(0.009)}$ & $10.6_{(2.33)}$ & $52.85_{(10.94)}$ \\
$(\frac{1}{3}, \frac{1}{12}, \frac{1}{48}, \frac{1}{192})$ & $14.1482_{(0.007)}$ & $6.8_{(0.75)}$ & $54.21_{(8.45)}$ \\
\hline
$(\frac{1}{6}, \frac{1}{12}, \frac{1}{24}, \frac{1}{48}, \frac{1}{96}, \frac{1}{192})$ & $14.1571_{(0.024)}$ & $11.0_{(1.90)}$ & $89.10_{(11.06)}$ \\
$\mathbf{(\frac{1}{6}, \frac{1}{12}, \frac{1}{24}, \frac{1}{48}, \frac{1}{96})}$ & $14.1733_{(0.022)}$ & $9.4_{(1.96)}$ & $68.51_{(7.92)}$ \\
$(\frac{1}{6}, \frac{1}{24}, \frac{1}{96})$ & $14.1692_{(0.008)}$ & $6.4_{(0.80)}$ & $56.73_{(1.86)}$ \\
$(\frac{1}{6}, \frac{1}{96})$ & $14.1499_{(0.034)}$ & $5.6_{(1.02)}$ & $58.46_{(5.42)}$ \\
\hline
$(\frac{1}{12}, \frac{1}{24}, \frac{1}{48}, \frac{1}{96}, \frac{1}{192})$ & $14.1782_{(0.010)}$ & $7.6_{(0.80)}$ & $119.25_{(3.44)}$ \\
$(\frac{1}{12}, \frac{1}{24}, \frac{1}{48}, \frac{1}{96})$ & $14.1614_{(0.012)}$ & $6.0_{(0.63)}$ & $96.11_{(2.67)}$ \\
$(\frac{1}{12}, \frac{1}{48}, \frac{1}{192})$ & $14.1599_{(0.007)}$ & $5.2_{(0.40)}$ & $107.43_{(4.80)}$ \\
\hline
 Halve every 1 loop & $14.1528_{(0.015)}$ & $5.0_{(0.00)}$ & $55.51_{(0.37)}$ \\
   Halve every 2 loops & $14.1660_{(0.017)}$ & $10.00_{(0.00)}$ & $71.41_{(1.92)}$ \\
 Halve every $1$ + Adaptive on $\bT^{(ex,B)}$ & $14.1506_{(0.016)}$ & $5.2_{(0.40)}$ & $59.27_{(2.47)}$ \\
Halve every $2$ + Adaptive on $\bT^{(ex,B)}$ & $14.1636_{(0.016)}$ & $9.8_{(0.75)}$ & $72.13_{(3.98)}$ \\ \hline
\end{tabular}
\caption{Tuning of the solver grid schedule $\{\bT^{(tr,b)}\}$ for the \texttt{M2} contract in Table~\ref{tab:contracts-tab}. See main text for explanation of the hybrid scheme in the  last two rows. The base choice is bolded. 
\label{tab:Stage3-RL-Grid-Schedule}}
\end{table}

It is not necessary to train on very fine solver grids: the stopping boundaries converge rapidly, so accurate American prices can be obtained without taking minuscule $\Delta t^{tr}$. We find that $\Delta t^{tr,B} = \tfrac{1}{150}$ is sufficient for typical stock volatilities. Table~\ref{tab:Stage3-RL-Grid-Schedule} shows that for the \texttt{M2} contract, although the ADNN is ultimately evaluated on $\Delta t^{ex} = \tfrac{1}{192}$, comparable prices are obtained by terminating the RL already at $\Delta t^{tr,B} = \tfrac{1}{96}$. Since RL loops on the finest grids are the most time-consuming (and most prone to instabilities that inflate loop counts), this also saves nearly 30\% of runtime.

For the initial grid size, 
a good rule of thumb is to use $N=20$ steps.
Using a coarser $\bT^{(tr,0)}$  increases the risk of concept drift and ultimately leads to lower rewards, as evidenced in Table~\ref{tab:Stage3-RL-Grid-Schedule}. Taking $N \gg 20$ steps is counterproductive: LS errors back-propagate, so finer discretization does not improve the Stage~1 data while adding computational cost.

In terms of grid refinement, our default is 
halve the step size at each transition. Decreasing the step size more rapidly, e.g., taking $\Delta t^{tr, b+1} = \tfrac{1}{4}\Delta t^{tr, b}$ or $\Delta t^{tr, b+1} = \tfrac{1}{8}\Delta t^{tr, b}$, decreases the overall number of learning loops~$L$. However,  as shown in Table~\ref{tab:Stage3-RL-Grid-Schedule} the runtime savings remain modest because the reduced loop count mainly shifts training away from computationally cheaper training on coarse grid to more expensive fine-grid RL loops. Aggressive schedules also increase concept drift, which makes NN instabilities more likely, and amplify their impact, since the algorithm has fewer chances to self-correct. As a result, $\Delta t^{tr, b} = 0.5^b \cdot \Delta t^{tr, 0}$ is the empirical best choice. 

To assess the effectiveness of the stopping-interval–based transitions introduced in Section~\ref{subsec:RL-Time}, we consider a version of our RL algorithm with a pre-specified number of learning loops per solver grid.  As shown in Table~\ref{tab:Stage3-RL-Grid-Schedule}, configurations with deterministic grid transition schedules perform worse, as some of the RL flexibility is removed.  As a further option, we implemented  a hybrid rule: deterministic transitions with a fixed number of $\ell \in [1,2]$ loops for intermediate solver grids, while  on the finest grid $\bT^{(tr, B)}$, we only conclude learning based on the adaptive rule in~\eqref{eq:SI}. This hybrid transition scheme performs on par with the fixed-loop variant. Taken together, these results indicate that the adaptive grid transitioning is effective and 
beats deterministic alternatives.

\begin{remark}
The learning rate $\eta^{[b]}$ directly impacts the magnitude of ADNN updates. Since solver grid refinement warrants progressively finer updates to the timing value estimates, $\eta^{[b]}$ should decrease as $\Delta t^{tr,b}$ shrinks. Smaller decay factors $\alpha_{\text{dec}}$ should therefore be used if the grid refinement schedule is more rapid than our default halving approach. 
\end{remark}

\subsection{Tuning Guidelines}\label{subsec:Tuning-Guidelines}

While CARLOS has many tuning parameters, most of them can be set to default values and need not be adjusted according to contract specifics. Based on our extensive experiments, we make the following recommendations. The overarching philosophy is to minimize computational effort without compromising learning stability. Table~\ref{tab:contracts-tab} then summarizes the contract-specific settings of the CARLOS algorithm based on the below guidelines. 
 
Stage~1 should employ a sufficiently large number of paths $K$ commensurate with the state dimension~$d$. As a baseline, we recommend setting $K = 10^4 \times d$. Larger $K$ is needed for deep OTM contracts, 
to ensure enough informative input-output pairs for ADNN initialization. 
Similarly, the number of input–output pairs $M$ should scale with the contract dimension $d$; we recommend $M = 10^4 \times d$. Although a larger~$M$ can improve the price estimate and reduce the standard error, the resulting gains are generally too modest to justify the additional computational cost. 

As a baseline, ReLU serves as a robust activation function. The ADNN hidden-layer width should scale with the option dimension~$d$. Specifically, we set the number of nodes to $\max\{30 \times d, 60\}$, implying that even one-dimensional contracts use 60 nodes. Finally, we recommend a batch size of 64, with the number of training epochs increasing for higher-dimensional options. For one- and two-dim.~contracts, we use 5 training epochs, increasing to 10 for $d \in \{3,4,5\}$, cf.~Table \ref{tab:Parameter-Setting}.

For the training density, we favor relatively high exploration initially, $\lambda_{\text{exl}}^{[0]} = 0.55, \lambda_{+}^{[0]} = \lambda_{-}^{[0]} = 0.2$ and $\lambda_{\text{ter}}^{[0]} = 0.05$ in \eqref{eq:KDE},
coupled with a gradual reallocation toward exploitation through $c_{\text{expl}}=0.25$ in Equation~\eqref{eq:lambda_update}. A moderately large number of validation paths~$V$ is needed to ensure stable grid transitions via the adaptive rule \eqref{eq:SI}. Note that larger $V$ makes the transition rule more strict (since the confidence interval shrinks), which  can ultimately reduce runtime by avoiding premature transitions to finer grids, where ADNN training is more computationally expensive, see Table \ref{tab:Stage3-RL-Pairs}. 

For the grid schedule, we start with roughly $N = 20$ grid steps in Stage~1 and halve the step size at each transition, up to around $\Delta t^{tr} \simeq \tfrac{1}{150}$, even if the exercise grid $\Delta t^{ex}$ is denser. Each grid transition makes further improvement of the timing value estimates roughly twice as computationally expensive, so there is a rapid saturation between price gains from further refinement relative to runtime. An initial learning rate $\eta^{[0]} = 10^{-4}$ is a good default, reduced by the decay factor $\alpha_{\text{dec}} = 0.7$ at each grid transition. This allows for larger ADNN updates on the coarser grids, while yielding more measured updates on the finer grids, where the timing value estimates are mainly refined near the stopping boundary.

The waiting period $\Delta^b_{\text{wait}}$ enforces exploration in the stopping region, and should decrease sublinearly as step size shrinks. 
To preserve sufficient exploration for updating timing value estimates inside $\cS^{[\ell]}$, we recommend $c_{\text{dlst}} = 1.3$ in \eqref{eq:Delta-Wait} assuming that $\Delta^{tr,b} = \Delta^{tr,0}/2^b$.

Ultimately, one may run the algorithm in the high-fidelity mode, where accuracy and stability is paramount. In that case, one should use a wide/deep network, a large $K,M,V$ and many epochs. If speed is equally important, the values in Table \ref{tab:Parameter-Setting}, used in our main benchmarking, offer good accuracy-runtime trade-offs and a guide for other contracts.

\begin{table}[ht]
\centering
{
\begin{tabular}{r|ccccc}
\hline
Parameter & \texttt{B1} & \texttt{B2} & \texttt{M2}a/b & \texttt{M3} & \texttt{M5}a/b \\
\hline\hline
Stage 1 paths $K$ & $10{,}000$ & $10{,}000$ & $20{,}000$ & $30{,}000$ & $50{,}000$ \\
NN Nodes $q$        & $60$ & $60$ & $60$ & $90$ & $150$ \\
NN Epochs              & $5$  & $10$  & $10$  & $10$  & $10$ \\
\hline
RL training inputs $M$        & $10{,}000$ & $20{,}000$ & $20{,}000$ & $30{,}000$ & $50{,}000$ \\
\hline
\end{tabular}}
\caption{Contract-specific parameter settings for the CARLOS algorithm, cf.~Table~\ref{tab:contracts-tab}. 
\label{tab:Parameter-Setting}}
\end{table}

\section{Conclusion}\label{sec:Conclude}

The developed CARLOS algorithm permits, for the first time, to solve continuous-time optimal stopping problems using a Monte Carlo simulation-based framework. To do so, we leveraged neural network techniques to train a single space-time aggregate surrogate that provides a stopping rule for any $t$. We then employed reinforcement learning logic that simultaneously trains this surrogate while gradually refining the exercise grid. Our main innovation is the iterative refinement which serves a dual purpose, ``killing two birds with one stone''. First, it integrates with the mini-batch-based NN training, while controlling for concept drift,  i.e., the shift in the timing-value distribution as the time grid changes. Second, it substantially lowers the running time compared to the brute force alternative of directly training a fine-grid ADNN. Instead, in CARLOS most of the training is done on coarser grids, achieving significant speed-up \emph{and} higher expected payoffs.  As we show, 5-12 RL loops are generally sufficient to numerically converge to the American-style formulation and provide an accurate continuous-time stopping boundary. Once RL training is complete, the ADNN can price American options on arbitrarily fine time grids, effectively enabling continuous-time stopping. For cases where PDE-based benchmarks are unavailable, CARLOS produces prices higher than all previously reported values in the literature.

Two other important innovations in CARLOS are the exploratory stopping and the adaptive training samples.
The continuation region expands as the grid is refined, motivating us to introduce a ``waiting interval'' for training inputs that start in the stopping region, in order to allow the ADNN to shift the stopping boundaries ``outward''. In addition, we prioritize ADNN training near the estimated stopping boundary---where even small timing-value updates matter---and near the contract maturity where timing values are very close to zero and it is important to stabilize them. To that end, we use a mix of exploratory, exploitative, and terminal inputs based on anchor sets. 

Looking ahead, two aspects of the method could be investigated further. First,  while we restrict the ADNN to fully connected feedforward networks, the optimal-stopping literature has explored alternative architectures, including convolutional NNs \cite{Felizardo}, Long Short-Term Memory (LSTM) networks \cite{Fathan}, recurrent NNs \cite{damera2023deep}, and randomized neural networks \cite{HerreraCalypso}. In particular, CNNs may be an appealing alternative for the ADNN because they are designed to recognize local patterns., which could help with adaptively refining the near-maturity region without substantially altering the learned timing-value estimates across the rest of the horizon.  
Second, additional ideas can be leveraged
 for constructing the training inputs, for instance exploiting contract specifics such as the ATM diagonal present in max-call options, or using batched training designs \cite{lyu2022adaptive}.

It would be worthwhile to generalize the idea of gradual and adaptive temporal grid refinement underlying CARLOS to related control problems, including multiple-stopping (swing option pricing) \cite{lauriere2025deep,deschatre2022deep}, optimal switching \cite{hu2020deep} and optimal impulse control \cite{ludkovski2022regression}. The approach should also prove fruitful for settings where time discretization is a limitation, such as deep BSDE solvers \cite{gao2023convergence}.

\bibliography{bb-ds-fellowship}
\bibliographystyle{siam}

\appendix
\renewcommand{\thefigure}{A.\arabic{figure}} 
\setcounter{figure}{0}
\renewcommand{\thetable}{A.\arabic{table}} 
\setcounter{table}{0}
\section{Supplementary Plots}

\begin{figure}[!ht]
    \centering

    \subfloat[]{
        \includegraphics[width=0.45\textwidth]{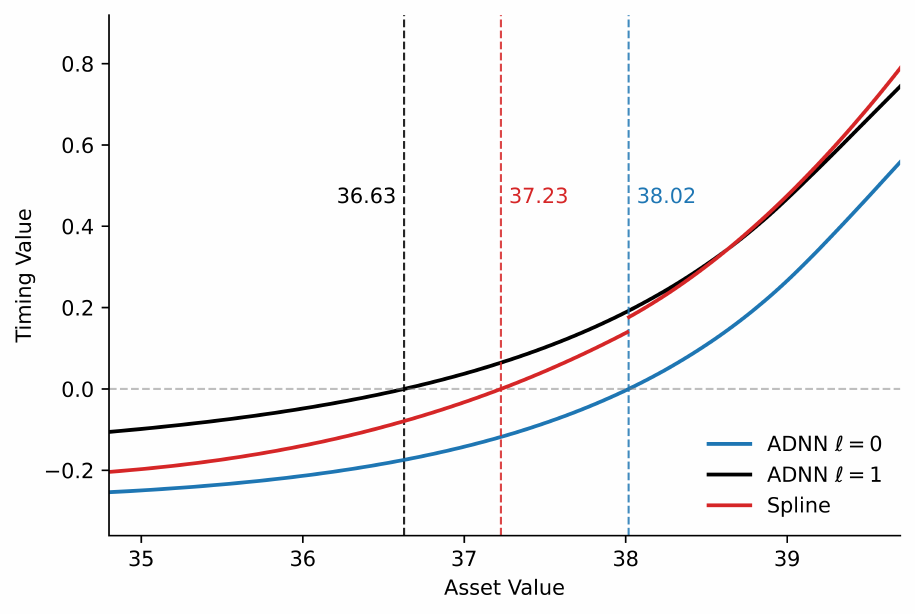}
        \label{fig:BP1-ADNN-Spline-Loop}
    }
    \hfill
    \subfloat[]{
        \includegraphics[width=0.45\textwidth]{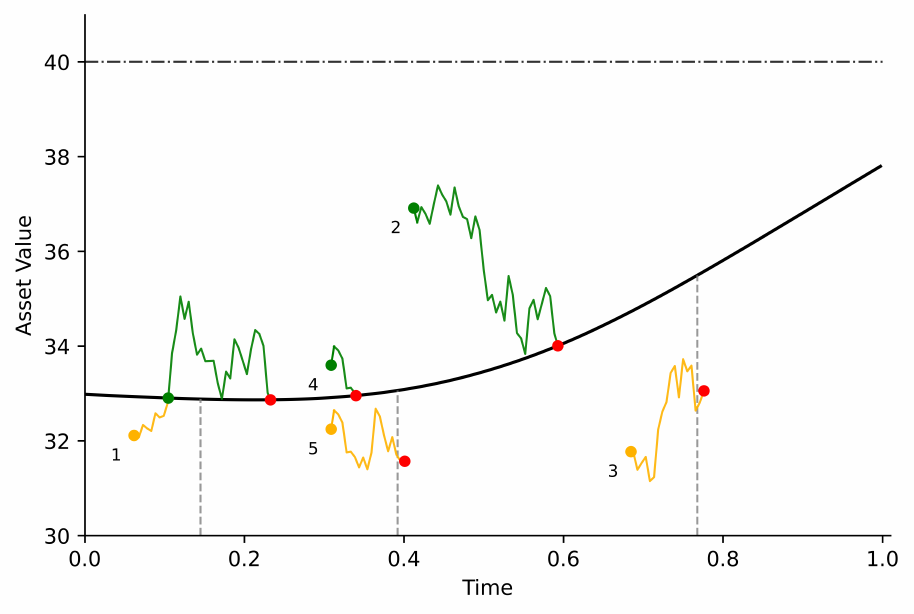}
        \label{fig:BP1-Paths-Color}
    }

    \caption{
    Exploratory stopping for the \texttt{B1} Put contract. \\
    (a) Timing values $\hat{\T}$ estimated at $t=0.89$ on $1{,}000$ discretized locations $x$ spanning $[30,40]$. The three curves show estimates from the ADNN $R^{[ 0]}$ at Stage~1 ($\ell = 0$, blue), the ADNN $R^{[1, 0]}$ after the first learning loop ($\ell = 1$, black), and a segmented cubic smoothing spline (red) fitted to Equation~(\ref{eq:RL-TV}) outputs. 
    To capture the piecewise structure of $y_{\text{dpf}}$ in \eqref{eq:Outputs-DPF}, two splines are fitted separately. Each output $y$ is the average timing value of $10{,}000$ simulated paths starting at $(t, x)$ with exercise frequency $\Delta t^{ex}=1/12$, subjected to a waiting period $\Delta_{\text{wait}}=0.065$, and stopped according to the ADNN $R^{[ 0]}$. Dashed vertical lines mark the zero-timing-value contours. \\ 
    (b) Sample paths (at finest frequency $\Delta t^{ex}=1/192$) illustrating the path labeling in Algorithm \ref{alg:Evaluation}. Yellow paths must exit the stopping region within a waiting period of $\Delta_{\text{wait}}=1/12$, indicated by vertical dashed gray lines. The stopping boundary is derived from the final ADNN after Stage~2 .
    }
    \label{fig:BP1-Outputs-Paths}
\end{figure}

\begin{figure}[!htb]
    \centering
    \subfloat[Paths initialized at $X_0$. Approximately 76.7\% of these paths are stopped. \label{fig:BP1-stop-time-distr-actual}]{
        \includegraphics[width=0.475\textwidth]{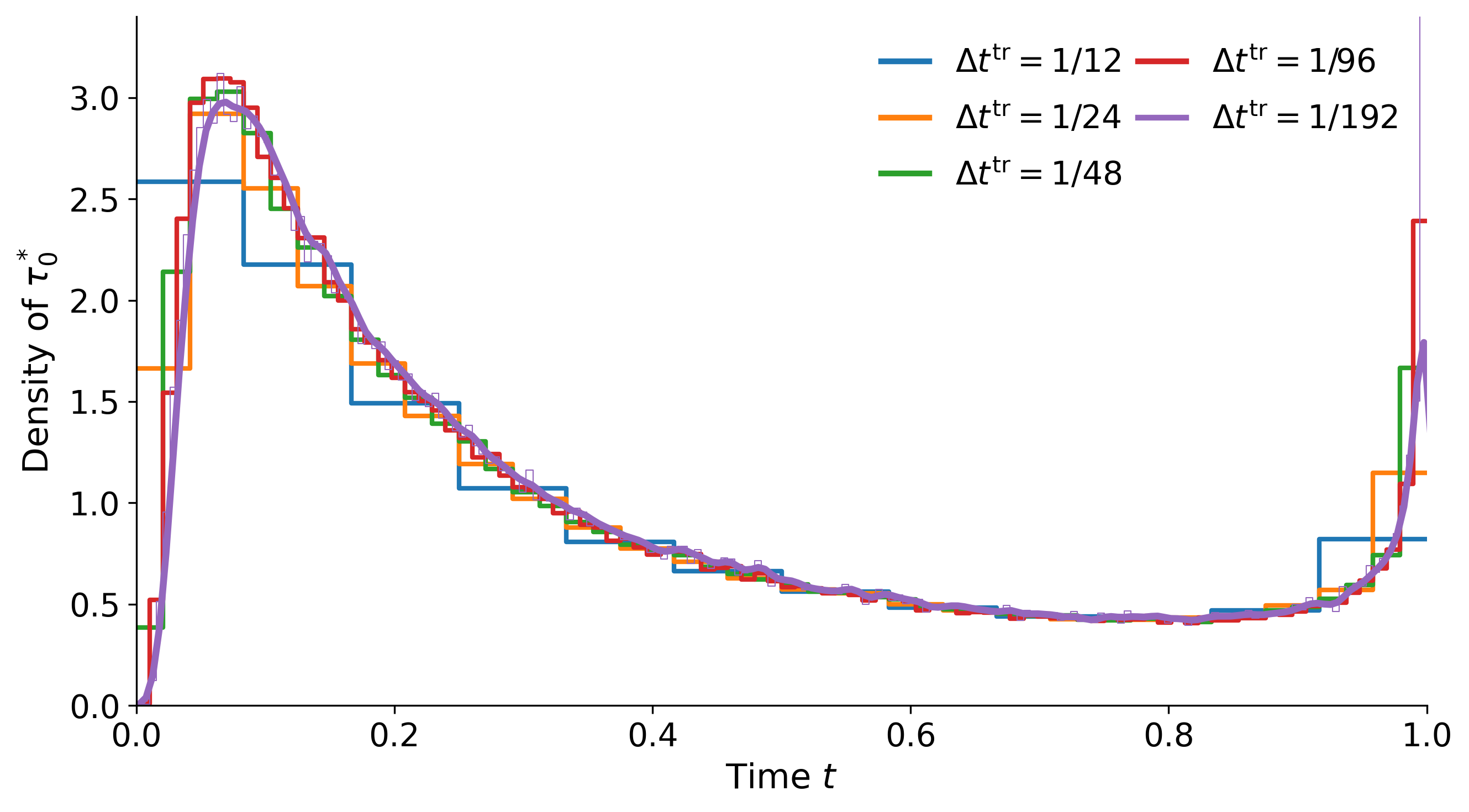}
    }
    \hfill
    \subfloat[Paths initialized at $x_0^p$, cf.~Equation~\eqref{eq:Init-Loc}. Approximately 65.1\% of these paths are stopped. \label{fig:BP1-stop-time-distr-jittered}]{
        \includegraphics[width=0.475\textwidth]{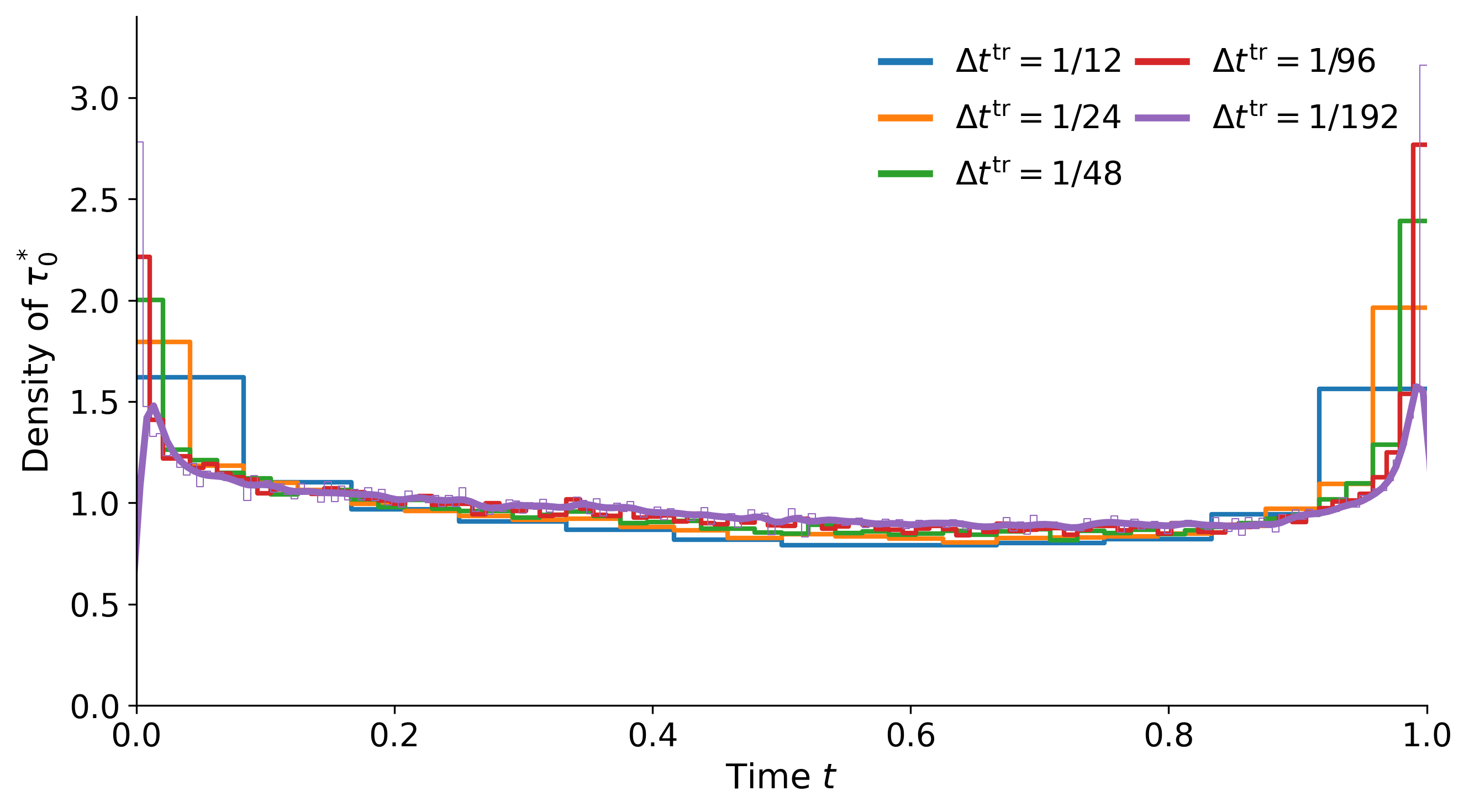}
    }

    \caption{Distributions of the optimal stopping time $\tau^*_0$ for the \texttt{B1} contract in Table~\ref{tab:contracts-tab} across PDE solver and exercise grids, estimated from $1.6 \times 10^6$ Monte Carlo paths.  
    The underlying stopping boundaries are computed with the Crank--Nicolson scheme with $\Delta t^{ex} = \Delta t^{\text{PDE}} = 1/192$ and $\Delta x = 0.02$.}
    \label{fig:tau-distribution}
\end{figure}

\clearpage

\section{Supplementary Tables}

\begin{table}[!ht]
\centering
\renewcommand{\arraystretch}{1.2}
\begin{tabular}{r|ccr}
\hline
\multicolumn{4}{c}{\textbf{Input-Output pairs} $M$} \\
\hline
& Price & Loops $L$ & Runtime \\
\hline\hline
$M=10{,}000$ & $14.1646_{(0.011)}$ & $8.8_{(1.94)}$ & $44.96_{(2.36)}$ \\
$M=20{,}000$ & $14.1714_{(0.015)}$ & $9.0_{(1.67)}$ & $57.00_{(5.11)}$ \\
$M=40{,}000$ & $14.1637_{(0.015)}$ & $8.0_{(0.63)}$ & $72.10_{(2.04)}$ \\
\hline
\multicolumn{4}{c}{\textbf{Validation Paths} $V$} \\
\hline
     $V=12{,}500$ & $14.1619_{(0.025)}$ & $8.4_{(1.02)}$ & $62.20_{(3.14)}$ \\
     $V=25{,}000$ & $14.1733_{(0.022)}$ & $9.4_{(1.96)}$ & $68.51_{(7.92)}$ \\
     $V=50{,}000$ & $14.1688_{(0.004)}$ & $8.2_{(2.48)}$ & $67.52_{(6.28)}$ \\
\hline
\multicolumn{4}{c}{\textbf{Exploration Window \eqref{eq:Delta-Wait}}}  \\
\hline
$c_{\text{dlst}}=0$ & $14.1573_{(0.012)}$ & $10.2_{(2.04)}$ & $58.45_{(6.79)}$ \\
\hline
$c_{\text{dlst}}=1.0$ & $14.1621_{(0.016)}$ & $9.2_{(2.40)}$ & $57.05_{(8.90)}$ \\
$c_{\text{dlst}}=1.1$ & $14.1686_{(0.020)}$ & $8.8_{(0.75)}$ & $56.43_{(2.67)}$ \\
$c_{\text{dlst}}=1.3$ & $14.1714_{(0.015)}$ & $9.0_{(1.67)}$ & $57.00_{(5.11)}$ \\
$c_{\text{dlst}}=1.5$ & $14.1524_{(0.023)}$ & $8.4_{(2.06)}$ & $53.08_{(4.88)}$ \\
\hline
$c_{\text{dlst}}=2$ & $14.1066_{(0.046)}$ & $9.2_{(2.93)}$ & $56.47_{(7.87)}$ \\
\hline
\end{tabular}
\caption{Tuning the number of input-output pairs $M$ per learning loop and the exploration power factor $c_{\text{dlst}}$ in Equation~(\ref{eq:Delta-Wait}) for the \texttt{M2} contract in Table~\ref{tab:contracts-tab}.}
\label{tab:Stage3-RL-Pairs}
\end{table}

\begin{table}[!ht]
\centering
\renewcommand{\arraystretch}{1.2}
\begin{tabular}{c|ccccc}
\hline
\multicolumn{6}{c}{\textbf{LSMC}} \\
\hline
$K$ & Stage~1 Price & Final Price & Loops $L$ & Stage 1 Time & Total Runtime \\
\hline\hline
$10{,}000$ & $14.0757_{(0.023)}$ & $14.1574_{(0.011)}$ & $9.4_{(1.96)}$ & $15.60_{(0.05)}$ & $42.44_{(4.21)}$ \\
$20{,}000$ & $13.9697_{(0.152)}$ & $14.1714_{(0.015)}$ & $9.0_{(1.67)}$ & $29.50_{(0.67)}$ & $57.00_{(5.11)}$ \\
$40{,}000$ & $13.7377_{(0.467)}$ & $14.1768_{(0.010)}$ & $9.6_{(1.02)}$ & $60.85_{(1.33)}$ & $88.07_{(1.70)}$ \\
\hline
\end{tabular}
\caption{Tuning the number of paths $K$ in Stage~1 for the \texttt{M2} contract in Table~\ref{tab:contracts-tab}. Stage~1 and Stage~2 prices are computed using $1.6\times10^6$ Monte Carlo paths at exercise frequency $\Delta t^{ex} = \frac{1}{192}$; the standard error of these estimates is approximately $0.013$. }
\label{tab:Stage1-LSMC} 
\end{table}

\begin{table}[!htb]
\centering
\begin{tabular}{cc|lllll|l}
    \multirow{2}{*}{Contract} & \multirow{2}{*}{\diagbox{$\Delta t^{tr}$}{$\Delta t^{ex}$}} & \multicolumn{5}{c|}{Monte Carlo Estimated Prices} & \multirow{2}{*}{PDE Price} \\ 
     &  &  $\sfrac{1}{12}$ & $\sfrac{1}{24}$ & $\sfrac{1}{48}$ & $\sfrac{1}{96}$ & $\sfrac{1}{192}$ &  \\
    \hline
    \multirow{5}{*}{\texttt{B2}} & $\sfrac{1}{12}$ & 1.459 & 1.467 & 1.469 & 1.468 & 1.466 & 1.455 \\
     & $\sfrac{1}{24}$ & 1.457 & 1.469 & 1.473 & $1.473^{**}$ & $1.473^{**}$ & 1.465 \\
     & $\sfrac{1}{48}$ & 1.454 & 1.468 & 1.474 & 1.475 & 1.476 & 1.471 \\
     & $\sfrac{1}{96}$ & 1.451 & 1.467 & 1.474 & 1.476 & 1.477 & 1.473 \\
     & $\sfrac{1}{192}$ & 1.448 & 1.465 & 1.473 & 1.476 & 1.478 & 1.475 \\
    \hline
    \multirow{5}{*}{\texttt{M2.A}} & $\sfrac{1}{12}$ & 14.137 & 14.172 & $14.177^{*}$ & 14.167 & 14.157 & 14.148 \\
     & $\sfrac{1}{24}$ & 14.132 & 14.181 & 14.196 & $14.198^{**}$ & 14.192 & 14.190 \\
     & $\sfrac{1}{48}$ & 14.111 & 14.174 & 14.201 & 14.209 & $14.211^{**}$ & 14.212 \\
     & $\sfrac{1}{96}$ & 14.096 & 14.168 & 14.200 & 14.210 & $14.214^{*}$ & 14.222 \\
     & $\sfrac{1}{192}$ & 14.080 & 14.157 & 14.192 & 14.209 & 14.214 & 14.228 \\
    \hline
    \multirow{5}{*}{\texttt{M2.B}} & $\sfrac{1}{12}$ & 15.718 & 15.738 & 15.747 & 15.741 &  15.734 & 15.742 \\
     & $\sfrac{1}{24}$ & 15.714 & 15.745 & 15.759 & $15.759^{**}$ & $15.756^{*}$ & 15.774 \\
     & $\sfrac{1}{48}$ & 15.705 & 15.744 & 15.761 & 15.769 & $15.768^{**}$ & 15.790 \\
     & $\sfrac{1}{96}$ & 15.692 & 15.738 & 15.762 & 15.773 & $15.776^{**}$ & 15.798 \\
     & $\sfrac{1}{192}$ & 15.682 & 15.733 & 15.760 & 15.771 & 15.777 & 15.802 \\
    \hline
\end{tabular}
\caption{Expected prices for the \texttt{B1}, \texttt{B2}, and \texttt{M2} options from Table~\ref{tab:contracts-tab}, estimated using $1.6 \times 10^6$ Monte Carlo simulated paths and PDE-based stopping rules. Row-wise price differences between adjacent $\Delta t^{ex}$ values are evaluated for statistical significance. If the difference between a price and its right neighbor is not statistically significant, the right-side price is marked with $\,^{*}$ at the 0.05 level and $\,^{**}$ at the 0.01 level. Standard deviations of the Monte Carlo estimated prices are approx. 0.0025 for \texttt{B1}, 0.0015 for \texttt{B2}, and 0.0115 for \texttt{M2}. For reference, the PDE-based prices for each $\Delta t^{tr}$ are also provided. The parameters for CN and explicit finite-difference methods are: \texttt{B1} ($\Delta t^{\text{PDE}} = 1/192$, $\Delta x = 0.02$), \texttt{B2} ($\Delta t^{\text{PDE}} = 10^{-5}$, $\Delta X^i = 0.2$ for $i = 1, 2$), and \texttt{M2} ($\Delta t^{\text{PDE}} = 10^{-5}$, $\Delta X^i = 0.8$ for $i = 1, 2$).}
\label{tab:PDE-tab-app} 
\end{table}

\section{Full Benchmark Descriptions}
\begin{table}[!ht]
\centering
\renewcommand{\arraystretch}{1.2}
\begin{tabular}{c||ccccc}
\multicolumn{6}{c}{\textbf{ADNN Parameter Settings}} \\
Parameter & \texttt{B1} & \texttt{B2} & \texttt{M2}a/b & \texttt{M3} & \texttt{M5}a/b \\
\hline
$K$ & $100{,}000$ & $100{,}000$ & $100{,}000$ & $200{,}000$ & $200{,}000$ \\
$\Delta t^{(N)}$ & 1/24 & 1/24 & 1/12 & 1/12 & 1/12 \\
\hline
Hidden nodes & $60$ & $60$ & $60$ & $90$ & $150$ \\
Epochs       & $5$  & $5$  & $5$  & $10$  & $10$ \\
\hline
\end{tabular}
\caption{Parameter configuration for the Stage~1 high-capacity ADNNs used to price the option contracts in Table~\ref{tab:contracts-tab}. Settings are grouped into LSMC parameters and ADNN architecture and training hyperparameters. All runs use ReLU activation and batch size of 64.}
\label{tab:Parameter-Stage-2}
\end{table}

\end{document}